\pdfoutput=1

\documentclass[11pt]{article}

\usepackage[preprint]{acl}

\usepackage{times}
\usepackage{latexsym}

\usepackage[T1]{fontenc}

\usepackage[utf8]{inputenc}

\usepackage{microtype}

\usepackage{inconsolata}

\usepackage[most]{tcolorbox}
\usepackage{graphicx}
\usepackage{pifont}
\usepackage{ulem}
\usepackage{enumitem}
\usepackage{xspace}
\usepackage{bbding}
\usepackage{multirow, booktabs}
\usepackage{makecell}
\usepackage{amssymb}
\usepackage{amsmath}
\usepackage{cleveref}
\usepackage{adjustbox}
\usepackage{listings}
\usepackage{float}
\usepackage{soul}
\usepackage{ulem}
\usepackage{twemojis}

\usepackage{nicematrix}

\usepackage{array}
\usepackage{collcell}

\makeatletter
\def\@fnsymbol#1{%
   \ifcase#1\or
   \TextOrMath \textdagger \dagger\or
   \TextOrMath \textdaggerdbl \ddagger \or
   \TextOrMath \textsection  \mathsection\or
   \TextOrMath \textparagraph \mathparagraph\or
   \TextOrMath \textbardbl \|\or
   \TextOrMath {\textdagger\textdagger}{\dagger\dagger}\or
   \TextOrMath {\textdaggerdbl\textdaggerdbl}{\ddagger\ddagger}\else
   \@ctrerr \fi
}
\makeatother

\usepackage{algorithm}
\usepackage[ruled,vlined,algo2e]{algorithm2e}
\usepackage{algpseudocode}

\usepackage{bbding}
\usepackage{fontawesome5}

\SetKwInOut{Input}{Input}\SetKwInOut{Output}{Output}
\newtheorem{definition}{Definition}

\definecolor{darksalmon}{rgb}{0.91, 0.59, 0.48}
\definecolor{emerald}{rgb}{0.31, 0.78, 0.47}
\definecolor{green(pigment)}{rgb}{0.0, 0.65, 0.31}
\definecolor{amaranth}{rgb}{0.9, 0.17, 0.31}
\definecolor{iris}{rgb}{0.35, 0.31, 0.81}
\definecolor{uu}{rgb}{0.95, 0.51, 0.51}
\definecolor{spirodiscoball}{rgb}{0.06, 0.75, 0.99}

\newcommand{\ourmethod}{{\fontfamily{lmtt}\selectfont \textbf{MasRouter}}\xspace}
\newcommand{\llmname}[1]{{\fontfamily{pcr}\selectfont {#1}}\xspace}

\title{MasRouter: Learning to Route LLMs for Multi-Agent Systems}

\author{
  \textbf{Yanwei Yue}{\footnotesize $^\bigstar$}
  \thanks{These authors contributed equally.} 
  \quad 
  \textbf{Guibin Zhang}{\footnotesize $^\bigstar$} 
  $^\dagger$
  \quad 
  \textbf{Boyang Liu}{\footnotesize $^\bigstar$} 
  $^\dagger$
  \quad 
  \textbf{Guancheng Wan}{\footnotesize $^\clubsuit$} \quad \\ 
  \textbf{Kun Wang}{\footnotesize $^\spadesuit$} \quad
  \textbf{Dawei Cheng}{\footnotesize $^\bigstar$} \quad 
  \textbf{Yiyan Qi}{\footnotesize $^{\blacklozenge}$} \\
  {\footnotesize $^\bigstar$}Tongji University 
  {\footnotesize $^\clubsuit$}Wuhan University  
  {\footnotesize $^\spadesuit$}Nanyang Technological University 
    {\footnotesize $^\blacklozenge$}IDEA\\
    {\faEnvelope} \textbf{Primary Contact:} \href{mailto:edwinmars77@gmail.com}{\texttt{edwinmars77@gmail.com}}
}

\begin{document}
\maketitle
\begin{abstract}
Multi-agent systems (MAS) powered by Large Language Models (LLMs) have been demonstrated to push the boundaries of LLM capabilities, yet they often incur significant costs and face challenges in dynamic LLM selection. Current LLM routing methods effectively reduce overhead in single-agent scenarios by customizing LLM selection for each query, but they overlook the critical decisions regarding collaboration modes and agent roles in MAS. In response to this challenge, we first introduce the problem of \textbf{Multi-Agent System Routing (MASR)}, which integrates all components of MAS into a unified routing framework. Toward this goal, we propose \ourmethod, the first high-performing, cost-effective, and inductive \textbf{MASR} solution. \ourmethod employs collaboration mode determination, role allocation, and LLM routing through a cascaded controller network, progressively constructing a MAS that balances effectiveness and efficiency. Extensive experiments demonstrate that \ourmethod is \textbf{(1) high-performing}, achieving a $1.8\%\sim8.2\%$ improvement over the state-of-the-art method on MBPP; \textbf{(2) economical}, reducing overhead by up to $52.07\%$ compared to SOTA methods on HumanEval; and \textbf{(3) plug-and-play}, seamlessly integrating with mainstream MAS frameworks, reducing overhead by $17.21\%\sim28.17\%$ via customized routing. The code is available at \url{https://github.com/yanweiyue/masrouter}.

\end{abstract}

\section{Introduction}
\begin{figure}[!t]
    \centering
    \includegraphics[width=1\linewidth]{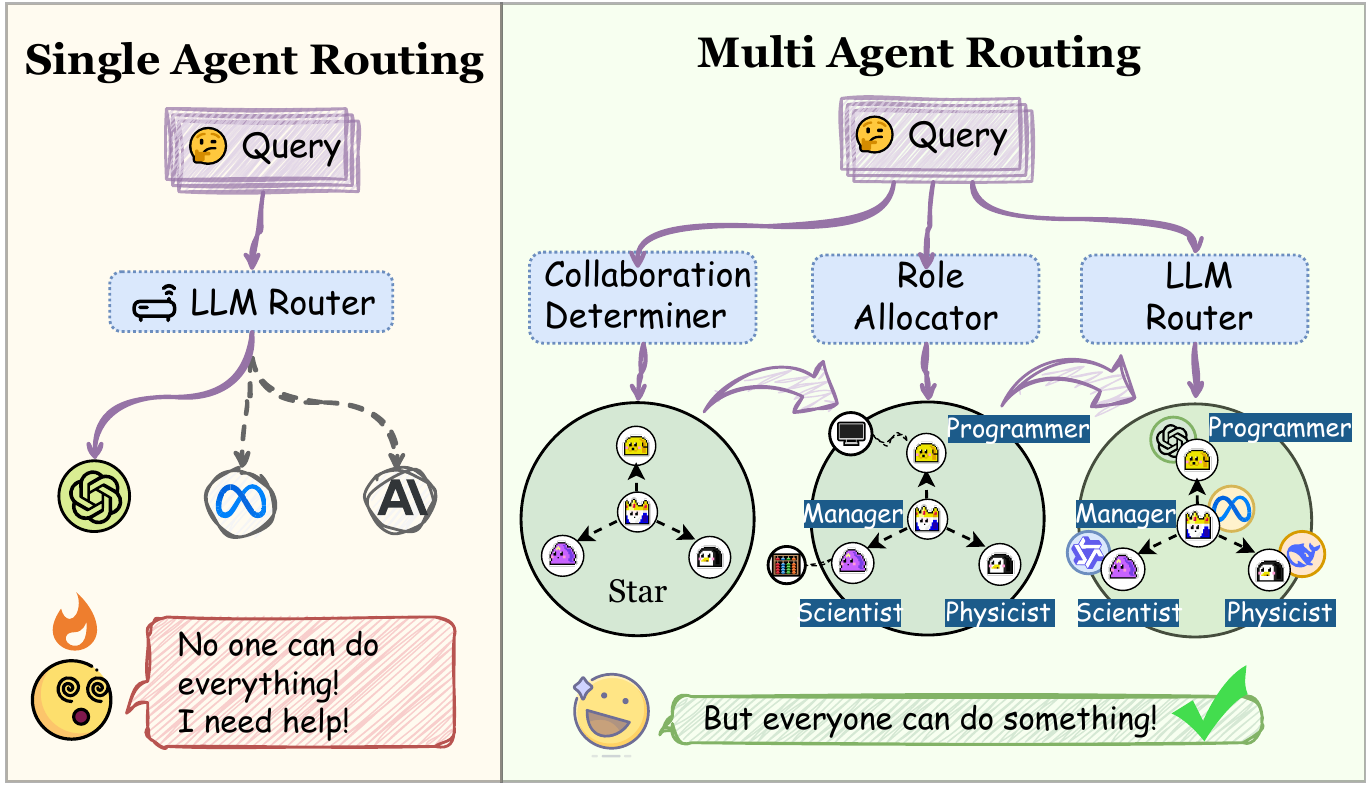}
    \vspace{-0.8cm}
    \caption{Paradigm comparison between single-agent routing and multi-agent routing.}
    \vspace{-0.4cm}
    \label{fig:intro}
\end{figure}

Recent advances in Large Language Model (LLM) based agents~\citep{generative-agents-simulacra,yao2023react,autogpt} have demonstrated remarkable success across various tasks, including code generation~\citep{autogen,guo2024deepseekcoderlargelanguagemodel}, mathematical reasoning~\citep{swan2023mathagentscomputationalinfrastructure,yu2024metamathbootstrapmathematicalquestions}, and embodied actions~\citep{voyager}. Building on the impressive capabilities of single agents, the LLM-based Multi-Agent System (MAS) has been proposed~\citep{generative-agents-simulacra,arXiv2023_MultiAgent-Debate,meta-gpt} to harness the collective intelligence and specialized expertise of multiple agents. 
As the pool of available LLMs continues to grow~\citep{chang2023surveyevaluationlargelanguage,minaee2024largelanguagemodelssurvey}, how to select the appropriate LLM to power single agents or MAS has increasingly captured the attention of the research community~\citep{hu2024routerbenchbenchmarkmultillmrouting, _akota_2024}. One might assume that, disregarding cost, choosing the largest LLM would always yield the best performance. 
However, larger LLMs have been shown to not always outperform their smaller counterparts~\citep{abdin2024phi3technicalreporthighly, lepagnol-etal-2024-small, shen2024smallllms}. At times, smaller LLMs can achieve superior performance at a lower computational cost.
Given this context, how to dynamically route the most appropriate LLM to empower an agent in a query-aware manner emerges as a highly compelling problem.

To address this, \textit{LLM routing} is proposed to intelligently assign the optimal LLM for each query~\citep{hu2024routerbenchbenchmarkmultillmrouting,srivatsa2024harnessingpowermultipleminds}. 
Early attempts employ a router, typically based on encoder models like BERT~\citep{devlin2019bertpretrainingdeepbidirectional}, to make a binary decision on whether to select a larger LLM~\citep{chen2023frugalgptuselargelanguage,ding2024hybridllmcostefficientqualityaware,ong2024routellmlearningroutellms}. 
More recent practices construct a router to assess the performance and cost of multiple LLMs, subsequently selecting the model that optimizes the trade-off between exploration and exploitation~\citep{dai2024costeffectiveonlinemultillmselection,mohammadshahi2024routoolearningroutelarge,feng2024graphroutergraphbasedrouterllm}. Although existing LLM routing methods~\citep{hu2024routerbenchbenchmarkmultillmrouting,stripelis-etal-2024-tensoropera} have been proven effective in directing the most appropriate LLM for input queries, they are limited to single-agent scenarios~\citep{cot,reflexion,agentgpt} yet not ready for MAS.
However, we argue that the availability of a routing method for MAS is even more essential. On the \textit{performance} perspective, multi-agent systems have been proven to outperform single-agent approaches significantly~\citep{chen2023agentverse,arXiv2023_MultiAgent-Cooperation}. On the \textit{cost} perspective, though impressive in performance, existing multi-agent pipelines inherently introduce substantial token overhead and increased economic costs~\citep{zhuge2024gptswarm}, which further necessitates the use of routing to mitigate overhead. In this context, there is a critical need to address this gap: \textit{How can we effectively route LLMs for MAS to balance performance and costs?}

A seemingly intuitive solution is to directly transfer single-agent LLM routing methods to MAS. However, MAS needs to select the appropriate agent collaboration modes~\citep{arXiv2023_MultiAgent-Debate,zhang_aflow_2024} for different tasks and establish a reasonable division of labor~\citep{meta-gpt,zhang2023exploring}, which previous routing methods fail to achieve. For example, for software development tasks, an ideal MAS routing method could design a hierarchical workflow with sequential steps such as requirements analysis, algorithm design, code development, and testing, each requiring corresponding role profiles~\citep{morethancode,zingg2023detectingoptimisingteaminteractions}. Against this backdrop, we argue that routing in MAS involves more tasks than just LLM recommendations: \ding{182} \textbf{Collaboration Mode Determination:} Choosing the optimal communication mechanisms (\textit{e.g.}, Chain~\citep{software-dev}, Tree~\citep{ishibashi2024selforganize-mother}, Graph~\citep{chatllm-network}) for varying task complexities. This involves identifying the most efficient and adaptable multi-agent topology~\citep{zhuge2024gptswarm,zhang2024cutcrapeconomicalcommunication} that minimizes overhead while ensuring flexibility and scalability in more complex scenarios. \ding{183} \textbf{Dynamic Agent Number:} Determining the number of expert agents required~\citep{huang2024hardertasksneedexperts,aghdam2024damoedynamicexpertallocation} based on the difficulty of the input. \ding{184} \textbf{Agent Role Allocation:} Selecting suitable role to the agent according to the query domain~\citep{chen2023agentverse,feng2024graphroutergraphbasedrouterllm} to ensure efficient task division, creating a system greater than the sum of its parts~\citep{shang2024agentsquareautomaticllmagent}.
\ding{185}\textbf{Agent LLM Routing:} Assigning each agent the appropriate LLM based on the collaborative topology and the role of each LLM~\citep{feng2024graphroutergraphbasedrouterllm}.

In light of the scrutinizing challenges, we for the first time introduce the concept of \textbf{LLM-based \uwave{M}ulti-\uwave{A}gent \uwave{S}ystem \uwave{R}outing (MASR):}

\noindent\fbox{%
    \parbox{0.95\linewidth}{%
       \textbf{Multi-Agent Systems Routing (MASR):} \textit{Given a pool of available LLMs, collaborative communication modes, and possible agent roles, an optimal MAS Router for any query $q$ should: (1) identify appropriate multi-agent collaboration modes, (2) allocate agent roles efficiently, and (3) assign the appropriate LLM to each agent, thereby balancing performance and cost.
       }
    }
}

The formal definition of MASR is provided in \Cref{sec:formalization}. To construct a router that ideally adheres to the MASR principles, we propose an effective, token-economical, and inductive LLM-powered \uwave{M}ulti-\uwave{A}gent \uwave{S}ystem \uwave{Router}, termed \ourmethod. Technically, \ourmethod integrates collaboration mode determiner, agent role allocator, and agent LLM router into a unified routing framework: \ding{182} \textbf{Collaboration determiner} employs a variational latent variable model to route the user query to a suitable collaboration module; \ding{183} \textbf{Role allocator} progressively generates agent roles through a structured probabilistic cascade; \ding{184} \textbf{LLM router} models the LLM backbone recommendation for each agent as a multinomial distribution problem. Ultimately, \ourmethod constructs a MAS that simultaneously \textbf{balances effectiveness and efficiency}.

Our contributions can be summarized as follows:
\begin{itemize}[leftmargin=*]
\vspace{-0.4em}\item \textbf{\textit{Problem Definition.}} We \textit{for the first time} formally define Multi-Agent System Routing (MASR), which specifies the requirements for MAS routing: assigning the appropriate collaboration mode, agent roles, and LLMs to each query, thereby improving response quality and reducing unnecessary overhead.
\vspace{-0.7em}\item \textbf{\textit{Practical Solution.}} We propose \ourmethod, a modular MASR solution that utilizes a cascaded controller network to construct a high-performing and resource-efficient MAS progressively. Besides, \ourmethod can seamlessly integrate with mainstream MAS to achieve efficient routing with significantly lower inference cost.
\vspace{-0.7em}\item \textbf{\textit{Experimental Validation.}} Extensive experiments across five benchmarks show that \ourmethod is: \textbf{(I) high-performing}, surpassing RouterDC, the state-of-the-art routing method, by $3.51\%$ on average; \textbf{(II) economical}, reducing the overhead on HumanEval from $0.363\$$ to $0.185\$$; \textbf{(III) inductive and plug-and-play}, generalizing to unseen LLM backbones and collaboration modes and seamlessly combining with mainstream multi-agent systems with $17\%\sim28\%\downarrow$ fewer cost.
\end{itemize}

\section{Related Work}
\vspace{-0.4em}
\paragraph {Multi-Agent System.}
Contemporary LLM-based multi-agent systems (MAS) can be broadly categorized into two paradigms: \textbf{(1) Fixed agentic networks} with pre-established, manually crafted architectures, from debate~\citep{chateval,debate2-thu}, collaboration~\citep{eot,NAACL2024_Agent-Self-Collaboration} to competetive~\citep{zhao2023competeai}. MacNet~\citep{qian2024scaling} systematically analyzed typical multi-agent collaboration topologies such as chain, tree, graph, \textit{etc.}; \textbf{(2) Dynamic agentic networks} that configure their structure and communication strategies based on real-time feedback and observations. ADAS~\citep{ADAS} and its follow-up works~\citep{zhang_aflow_2024,shang2024agentsquareautomaticllmagent} leverage search methods such as Monte Carlo Tree Search (MCTS) and evolutionary algorithm~\citep{zhang2025evoflow} to discover effective agent strategies. Other works like DyLAN~\citep{arXiv2023_Dynamic-LLM-Agent}, GPTSwarm~\citep{zhuge2024gptswarm} and AgentPrune~\citep{zhang2024cutcrapeconomicalcommunication} dynamically optimize the inter-agent topologies. Nevertheless, contemporary MAS is often LLM-homogeneous, \textit{i.e.}, relying exclusively on the same LLM backbone, failing to collectively organize heterogenous LLM-agents.

\vspace{-0.4em}
\paragraph{Single LLM Routing}
Efficient routing strategies for single LLMs have been extensively explored to balance computational cost and model performance.
Early attempts on LLM routing include HybridLLM~\citep{ding2024hybridllmcostefficientqualityaware}, RouteLLM~\citep{ong2024routellmlearningroutellms} and FrugalGPT~\citep{chen2023frugalgptuselargelanguage}, which primarily focus on binary routing and leverage techniques like sequential pipelines or preference-driven routing to enhance decision-making performance. More recent practices, including Rootoo~\cite{dai2024costeffectiveonlinemultillmselection}, C2MAB-V~\cite{mohammadshahi2024routoolearningroutelarge}, GraphRouter~\citep{feng2024graphroutergraphbasedrouterllm} and RouterDC~\cite{chen2024routerdc} have shifted towards multi-choice selection frameworks. However, existing routing methodologies mainly focus on single-agent scenarios, and their unawareness of inter-agent topology constrains their applicability to more complex tasks and limits scalability in larger systems.

\section{Formalization}
\label{sec:formalization}

\begin{figure*}[ht]
\centering
\includegraphics[width=1.0\linewidth]{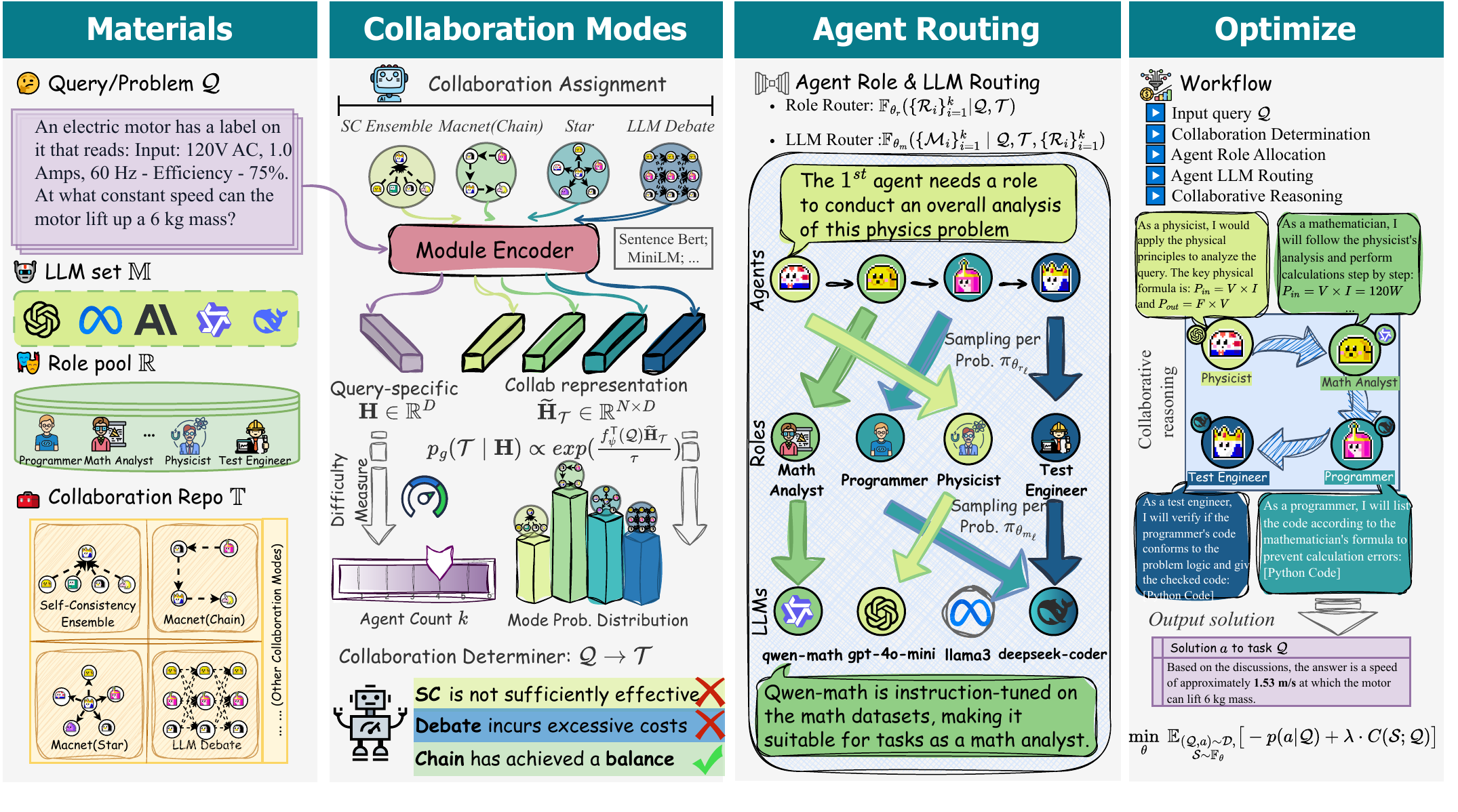}
\vspace{-2em}
\caption{The overall framework of our proposed \ourmethod.
}
\vspace{-0.5em}
\label{fig:framework}
\end{figure*}

In this section, we formalize our proposed LLM-based \textbf{Multi-Agent System Routing (MASR)} and introduce its optimization objectives.

\subsection{Notation Establishment}
\paragraph{Search Space.} We first define the search space of a multi-agent system as $\mathbb{S}=( \mathbb{M}, \mathbb{R}, \mathbb{T})$, where $\mathbb{M}$ denotes the pool of $N_m$ available LLM backbones, $\mathbb{R}$ represents the set of $N_r$ predefined agent roles (\textit{e.g.}, {analyst}, {programmer}, and {tester}), and $\mathbb{T}$ denotes the set of $N_t$ collaboration modes, including structures like Chain, Tree, LLM-Debate, \textit{etc}. Within the search space, a multi-agent system instance is defined as follows:

\begin{definition}[\textbf{Multi-Agent System}]
\label{def:mas}
The MAS $\mathcal{S}$ combines several specific LLM-powered agents with distinct identities, working together collaboratively:
\begin{equation}\label{eq:mas}
\begin{gathered}
\mathcal{S} = \{\{\mathcal{M}_i\}_{i=1}^k, \{\mathcal{R}_i\}_{i=1}^k,\mathcal{T}\},\\
\mathcal{M}_i \in \mathbb{M}, \mathcal{R}_i \in \mathbb{R}, \mathcal{T} \in \mathbb{T},\\
\end{gathered}
\end{equation}
where $\mathcal{M}$ corresponds to the selected LLM backbones and, similarly, $\mathcal{R}$ and $\mathcal{T}$ represent the chosen role and collaboration mode respectively. $k$ denotes the number of the LLM agents.
\end{definition}

\subsection{Definition of MASR}
Based on the MAS defined above, we formalize Multi-Agent Systems Routing (MASR):
\begin{definition}[\textbf{MASR}]
\label{def:masr}
MASR can be represented by the mapping function $f$, which maps from \(\mathbb{S}=(\mathbb{M}, \mathbb{R}, \mathbb{T})\) to a MAS $\mathcal{S}$ tailored for the query \( \mathcal{Q} \):
\vspace{-0.5em}
\begin{equation}\label{eq:masr}
\begin{gathered}
f : \mathbb{M} \times \mathbb{R} \times \mathbb{T} \xrightarrow{} \mathcal{S},\\
\pi\left( \mathcal{S}\right)=\mathbb{P}\left(\left\{\{\mathcal{M}_i\}_{i=1}^k,\{\mathcal{R}_i\}_{i=1}^k,\mathcal{T}\right\}\mid\mathcal{Q}\right),\\
\mathcal{M}_i\in\mathbb{M},\mathcal{R}_i\in\mathbb{R},\mathcal{T}\in\mathbb{T},S \subset \mathbb{S}\\
\end{gathered}
\end{equation}
where $\pi\left( \mathcal{S}\right)$ represents the probability of selecting multi-agent system $S$, conditioned on $\mathcal{Q}$.
\end{definition}
\paragraph{Optimization Objective.} Given a benchmark \(\mathcal{D}\) consisting of multiple queries \(\mathcal{Q}\) and corresponding oracle answers \(a\), the ideal \textbf{MASR} aims to optimize a strategy to jointly balance performance and cost. The objective function is defined as: 
\begin{equation}
\small
\underset{\mathbb{P}(\mathcal{S} | \mathcal{Q})}{\max}\; \mathbb{E}_{\substack{(\mathcal{Q},a) \sim \mathcal{D}, \\\mathcal{S}\in \mathbb{S} \sim \mathbb{P}(\mathcal{S} | \mathcal{Q})}} \big[ \underbrace{U(\mathcal{S}; \mathcal{Q}, a)}_{Utility} - \lambda \cdot \underbrace{C(\mathcal{S}; \mathcal{Q})}_{Cost} \big],
\end{equation}
where \(\mathbb{P}(\mathcal{S} | \mathcal{Q})\) represents the probability distribution of  \(\mathcal{S}\) conditioned on \(\mathcal{Q}\). The utility term $U(\mathcal{S}; q, a)$ measures the performance of the MAS, while the cost term $C(\mathcal{S}; q)$ quantifies the expected cost (\textit{e.g.}, LLM calls, API cost, token cost). $\lambda$ is a trade-off parameter.

\section{MasRouter}
\vspace{-0.6em}
\Cref{fig:framework} illustrates the overall framework of \ourmethod.
For a given query, \ourmethod samples the customized components of the MAS, through the collaboration determiner ($\triangleright$ \Cref{sec:methodcollab}), role allocator ($\triangleright$ \Cref{sec:methodrole}), and agent LLM router ($\triangleright$ \Cref{sec:methodllm}), together forming a task-adaptive MAS $\mathcal{S}$. After executing the sampled MAS, \ourmethod jointly optimizes the parameters of each selection module based on the performance/cost feedback ($\triangleright$ \Cref{sec:methodoptim}).

\subsection{Collaboration Mode Determination}
\label{sec:methodcollab}
Given a query \(\mathcal{Q}\), the core objective of \ourmethod is to customize an appropriate MAS from the search space $\mathbb{S}$ based on the complexity and domain of the query, thereby generating a sufficiently high-quality response:
\begin{equation}\label{eq:core}
p(a | \mathcal{Q}) = \int \mathcal{O}(a | \mathcal{S}) \; \mathbb{F}_\theta(\mathcal{S} | \mathcal{Q}) \, d\mathcal{S},\mathcal{S}\subset\mathbb{S},
\end{equation}
where \( \mathbb{F}_{\theta} \) represents the controller network parameterized by \( \theta \), which takes \( \mathcal{Q} \) and computes the underlying distribution of \( \mathcal{S} \). \( \mathcal{O}(\cdot |\cdot) \) denotes the conditional likelihood of obtaining the solution $a$ by executing $\mathcal{S}$. $\mathbb{F}_{\theta}$ is formulated as follows:
\vspace{-0.3em}
\begin{equation}
\mathbb{F}_\theta =\mathbb{F}_{\theta_m}\circ\mathbb{F}_{\theta_r}\circ\mathbb{F}_{\theta_t},
\end{equation}
where \( \mathbb{F}_{\theta_t}:\mathcal{Q}\rightarrow\mathcal{T} \) is the collaboration mode determiner, \( \mathbb{F}_{\theta_r}:(\mathcal{Q},\mathcal{T})\rightarrow\{\mathcal{R}_i\}_{i=1}^{k}\) denotes the role allocator, and \( \mathbb{F}_{\theta_m}:(\mathcal{Q},\mathcal{T},\{\mathcal{R}_i\}_{i=1}^{k})\rightarrow\{\mathcal{M}_i\}_{i=1}^{k} \) represent the LLM backbone router. Inspired by human collaboration~\citep{humangroup,chen2023agentverse}, \ourmethod first constructs a team management framework for the MAS using $\mathbb{F}_{\theta_t}$, then recruits suitable talents and defines the division of tasks using $\mathbb{F}_{\theta_r}$, and finally endows each agent with the unique intelligence through $\mathbb{F}_{\theta_m}$. For \( \mathbb{F}_{\theta_t} \), since the relationship between collaborative modes and  queries is generally difficult to characterize explicitly, we employ a variational latent model to capture their underlying semantic associations: 
\begin{equation}
\label{eq:collabaration}
\small
\begin{aligned}
\mathbb{F}_{\theta_t}(\mathcal{T}\mid \mathcal{Q})
= \int p_g(\mathcal{T} \mid \mathbf{H}) \, p_h(\mathbf{H} \mid \mathcal{Q}) \, d\mathbf{H},\mathcal{T}\in\mathbb{T}
\end{aligned}
\end{equation}
where \( p_h(\mathbf{H} | \mathcal{Q}) \) denotes the prior probability distribution of the latent representation, and \( p_g(\mathcal{T} \mid \mathbf{H}) \) decodes the probability of collaborative patterns, implemented as follows:
\begin{equation}
\label{eq:endecoder}
\begin{gathered}
p_h(\mathbf{H} \mid \mathcal{Q}) = \mathcal{N}(\mathbf{H}; \mu_t(\mathcal{Q}), \text{diag}(\sigma_t^2(\mathcal{Q}))),\\
p_g(\mathcal{T} \mid \mathbf{H}) \propto exp(\frac{f_{\psi}^{\mathsf{T}}(\mathcal{Q})\widetilde{\mathbf{H}}_{\mathcal{T}}}{\tau}),\tau >0
\end{gathered}
\end{equation}
where \(\mu_t(\cdot)\) and \(\sigma^2_t(\cdot)\) obtains the mean and variance of \(\mathbf{H}\), respectively, and \(\widetilde{\mathbf{H}}_{\mathcal{T}} = g_{\phi}(f_{\psi}(\mathcal{T}), \mathbf{H})\) embeds the relationship between the query and the collaborative patterns into the latent space. Here, \(f_{\psi}:\mathcal{Q}\rightarrow\mathbb{R}^D\) is a text encoder (\textit{e.g.}, Sentence-BERT~\citep{reimers2019sentence}, MiniLM~\cite{wang2020minilm}) that extracts the semantic information of the query. \(g_{\phi}:\mathbb{R}^D\times\mathbb{R}^D\rightarrow\mathbb{R}^D\) produces the refined representation of the candidate \(\mathcal{T}\).

With $\mathbb{F}_{\theta_t}$, we have now customized the collaborative patterns for \(\mathcal{Q}\). Nevertheless, the number of agents utilized in $\mathcal{T}$ remains undetermined. We leverage the hidden embedding of the query to derive the number of agents \(k = \left\lceil \delta(\mathbf{H}) \cdot \gamma \right\rceil\), where \(\delta: \mathbb{R}^D \rightarrow [0,1]\) is a learnable complexity mapping function, and \(\gamma\) is the hyperparameter representing the maximum number of agents.
\vspace{-0.3em}
\subsection{Agent Role Allocation}
\label{sec:methodrole}
For a MAS \(\mathcal{S} = \{\{\mathcal{M}_i\}_{i=1}^k, \{\mathcal{R}_i\}_{i=1}^k,\mathcal{T}\}\), \Cref{sec:methodcollab} has specified \(\mathcal{T}\) and \(k\). Subsequently, we will assign the appropriate role $\mathcal{R}_i$ to each agent in $\mathcal{S}$. Roles among different agents often have a sequential order and interdependencies. For example, we first need a programmer to write the code, and then a test engineer to validate and debug it. Correspondingly, the \textbf{role allocator} $\mathbb{F}_{\theta_r}$ formalizes role generation through a structured probabilistic cascade: 
\begin{equation}
\small
\mathbb{F}_{\theta_r}(\{\mathcal{R}_i\}_{i=1}^k|\mathcal{Q},\mathcal{T}) 
= \prod_{\ell=1}^k \pi_{r_\ell}(\mathcal{R}_\ell | \mathcal{Q}, \{\mathcal{R}_j\}_{j=1}^{\ell-1},\mathcal{T}),
\end{equation}
where $\pi_{r_\ell}$ denotes the probability of generating the $\ell$-th role is based on $\mathcal{Q}$, the selected $\mathcal{T}$, and the prior $\ell-1$ role profiles. We iteratively compute it as follows:
\begin{equation}
\small
\begin{gathered}
\label{eq:roledecoder}
\pi_{r_\ell}(\mathcal{R}_\ell | \mathcal{Q},\mathcal{T}, \{\mathcal{R}_j\}_{j=1}^{\ell-1})\propto exp(\frac{\mathbf{H}_{\mathcal{R}_{\ell-1}}^\mathsf{T}\widetilde{\mathbf{H}}_{\mathcal{R}_\ell}}{\tau}),\\
R_\ell\in\mathbb{R}, \tau>0,
\end{gathered}
\end{equation}
where $\mathbf{H}_{\mathcal{R}_{\ell-1}} = \operatorname{FFN}(\mathbf{H}\| \widetilde{\mathbf{H}}_\mathcal{T}\| \frac{\sum_{j=1}^{\ell-1}\widetilde{\mathbf{H}}_{\mathcal{R}_j}}{\ell-1})$ denotes the implicit representation of the accumulated semantics from the role allocation process of the first \(\ell - 1\) roles under the \(\mathcal{Q}\) and \(\mathcal{T}\). $\widetilde{\mathbf{H}}_{\mathcal{R}_\ell} = g_\phi(f_{\psi}(\mathcal{R}_\ell),\mathbf{H}_{\mathcal{R}_{\ell-1}})$ captures the dynamic features exhibited by the current candidate role within the context of the previously assigned roles. Through \Cref{eq:roledecoder}, \ourmethod progressively determines the roles for all agents in $\mathcal{S}$. Afterward, the remaining task is to provide each agent with its driving force by routing to an appropriate LLM backbone.

\subsection{Agent LLM Routing}
\label{sec:methodllm}
Each LLMs have its own strengths and drawbacks~\citep{barandoni2024automatingcustomerneedsanalysis}, and the goal of LLM routing is to leverage their unique capabilities. For example, for mathematical problems, we would opt for an LLM that is particularly proficient in mathematics or one that has been specifically fine-tuned. Therefore, we posit that assigning an LLM to a specific agent primarily depends on the task's domain and difficulty, as well as their corresponding role. We implement $\mathbb{F}_{\theta_m}$ by computing the probability of selecting $\mathcal{M}_i$ based on the query and the preceding role routing. It then views the process of LLM routing for multiple agents as a multinomial distribution problem:
\begin{equation}\label{eq:llm-selection}
\small
\begin{gathered}
\mathbb{F}_{\theta_m}(\{\mathcal{M}_i\}_{i=1}^k \mid \mathcal{Q}, \mathcal{T},\{\mathcal{R}_i\}_{i=1}^k)\\
= \tbinom{k}{n_1,n_2,...,n_{N_m}}\cdot \prod_{
\ell=1}^{N_m}\pi_{m}^{n_\ell}(\mathcal{M}_{\ell}\mid\mathcal{Q},\mathcal{T},\{\mathcal{R}_i\}_{i=1}^k),
\end{gathered}
\end{equation}
where \(\binom{k}{n_1, n_2, \dots, n_{N_m}} = \frac{k!}{n_1! n_2! \dots n_{N_m}!}\) is the multinomial coefficient. It represents the number of ways to assign \( k \) agents to \( N_m \) different LLMs, with the \( i \)-th LLM selected \( n_i \) times. \(\pi_m\) denotes the probability of each LLM being selected in the global context:
\begin{equation}
\small
\begin{aligned}
\pi_{m}(\mathcal{M}_\ell | \mathcal{Q}, \mathcal{T}, \{\mathcal{R}_i\}_{i=1}^k) \propto exp(\frac{\mathbf{H}^\mathsf{T}_{\mathcal{M}}\widetilde{\mathbf{H}}_{\mathcal{M}_\ell}}{\tau}),
\end{aligned}
\end{equation}
where \(\mathbf{H}_{\mathcal{M}} = \operatorname{FFN}(\mathbf{H} \| \widetilde{\mathbf{H}}_\mathcal{T} \| \frac{\sum_{j=1}^{k}\widetilde{\mathbf{H}}_{j}}{k})\) aggregates the embedding of the query, collaborative patterns, and selected roles. \(\widetilde{\mathbf{H}}_{\mathcal{M}_\ell} = g_\phi(f_{\psi}(\mathcal{M}_\ell), \mathbf{H}_{\mathcal{M}})\) computes the latent representation of each LLM. Based on \(\mathbf{H}_{\mathcal{M}}\) and \(\widetilde{\mathbf{H}}_{\mathcal{M}_\ell}\), the compatibility between each LLM and the constructed system is obtained, which is proportional to the probability of selecting $M_\ell$.

As stated above, we have customized a MAS for the query. Only one final hurdle is left before achieving end-to-end training: the number of agents \( k \) becomes non-differentiable due to the rounding operation. To ensure smooth gradient flow, we replace \( k \) with its pre-rounded floating-point value and approximate the multinomial coefficient in \Cref{eq:llm-selection} as follows:
\begin{equation}
\small
\tbinom{k}{n_1,n_2,...,n_{N_M}} \approx \frac{\Gamma(\delta(\mathbf{H}) \cdot \gamma+1)}{\Gamma(n_1+1)\Gamma(n_2+1)...\Gamma(n_{N_M}+1)},
\end{equation}
where $\Gamma(\cdot)$ denotes the Gamma function.

\begin{table*}[!t]
\setlength{\tabcolsep}{10pt}
\resizebox{\linewidth}{!}{
\begin{tabular}{l|c|cc|cccccc}
\Xhline{1.2pt}
\rowcolor{CadetBlue!20} 
{\textbf{Method}}&\textbf{LLM} & \textbf{Mul.} & \textbf{Rout.} & \textbf{MMLU} & \textbf{GSM8K} & \textbf{MATH} & \textbf{HumanEval} & \textbf{MBPP} & {\textbf{Avg.}} \\
\Xhline{1.2pt}

\rowcolor{gray!10}&  \llmname{gpt-3.5-turbo} & \textcolor{darksalmon}{\XSolidBrush} & \textcolor{darksalmon}{\XSolidBrush} & 69.28 & 77.97 & 44.12 & 72.05 & 70.20 & 66.72 \\
\rowcolor{gray!10}& \llmname{gpt-4o-mini} & \textcolor{darksalmon}{\XSolidBrush} & \textcolor{darksalmon}{\XSolidBrush} & 77.81 & 93.17 & 66.09 & 85.71 & 72.20 & 79.00 \\
\rowcolor{gray!10}&  \llmname{claude-3.5-haiku} & \textcolor{darksalmon}{\XSolidBrush} & \textcolor{darksalmon}{\XSolidBrush} & 67.97 & 92.16 & 65.89 & 86.33 & 73.40 & 77.15 \\
\rowcolor{gray!10}& \llmname{gemini-1.5-flash} & \textcolor{darksalmon}{\XSolidBrush} & \textcolor{darksalmon}{\XSolidBrush} & 80.04 & 92.67 & 74.39 & 82.61 & 73.00 & 80.54 \\
\rowcolor{gray!10} \multirow{-5}{*}{Vanilla} &  \llmname{llama-3.1-70b} & \textcolor{darksalmon}{\XSolidBrush} & \textcolor{darksalmon}{\XSolidBrush} & 79.08 & 92.68 & 60.31 & 80.75 & 68.20 & 76.20 \\
\hline

 & \llmname{gpt-4o-mini} & \textcolor{darksalmon}{\XSolidBrush}  & \textcolor{darksalmon}{\XSolidBrush} & 78.43 & 93.68 & 67.24 & 86.69 & 69.60 & 79.13 \\

\multirow{-2}{*}{ CoT~\citep{cot}} & \llmname{gemini-1.5-flash} & \textcolor{darksalmon}{\XSolidBrush}  & \textcolor{darksalmon}{\XSolidBrush} & 81.35 & 92.92 & 74.34 & 81.37 & 73.00 & 80.60\\

& \llmname{gpt-4o-mini} & \textcolor{darksalmon}{\XSolidBrush}  & \textcolor{darksalmon}{\XSolidBrush} & 81.05 & 93.43 & 67.05 & 87.58 & 75.80 & 80.98 \\

\multirow{-2}{*}{ComplexCoT~\cite{fu2022complexity}} &  \llmname{gemini-1.5-flash} & \textcolor{darksalmon}{\XSolidBrush}  & \textcolor{darksalmon}{\XSolidBrush} & 80.74 & 92.01 & 75.11 & 80.12 & 71.80 & 79.96\\

 & \llmname{gpt-4o-mini} & \textcolor{darksalmon}{\XSolidBrush}  & \textcolor{darksalmon}{\XSolidBrush} & 81.05 & 93.32 & 66.28 & 87.58 & 73.00 & 80.25 \\

\multirow{-2}{*}{SC(CoT)~\cite{wang2023selfconsistency}} & \llmname{gemini-1.5-flash} & \textcolor{darksalmon}{\XSolidBrush}  & \textcolor{darksalmon}{\XSolidBrush} & 81.66 & 93.43 & 74.37 & 80.75 & 72.00 & 80.44 \\

& \llmname{gpt-4o-mini} & \textcolor{darksalmon}{\XSolidBrush}  & \textcolor{darksalmon}{\XSolidBrush} & 82.35 & 93.94 & 66.86 & 88.19 & 75.80 & 81.43 \\

\multirow{-2}{*}{SC(ComplexCoT)~\citep{wang2023selfconsistency}} &  \llmname{gemini-1.5-flash} & \textcolor{darksalmon}{\XSolidBrush}  & \textcolor{darksalmon}{\XSolidBrush} & 82.39 & 92.98 & \underline{75.31} & 81.99 & 73.60 & 81.25 \\
\hline

\rowcolor{gray!10}  & \llmname{gpt-4o-mini} &  \textcolor{green(pigment)}{\Checkmark} & \textcolor{darksalmon}{\XSolidBrush} & 82.01 & 94.40 & 64.72 & 85.63 & 75.40 & 80.43 \\

\rowcolor{gray!10} \multirow{-2}{*}{\cellcolor{gray!10} Chain~\cite{qian2024scaling}} &  \llmname{gemini-1.5-flash} & \textcolor{green(pigment)}{\Checkmark} & \textcolor{darksalmon}{\XSolidBrush} & 83.01 & 93.13 & 72.10 & 82.50 & 73.20 & 80.79 \\

\rowcolor{gray!10} & \llmname{gpt-4o-mini} & \textcolor{green(pigment)}{\Checkmark} & \textcolor{darksalmon}{\XSolidBrush} & 82.98 & 93.89 & 65.11 & 87.50 & 75.60 & 81.02 \\

\rowcolor{gray!10} \multirow{-2}{*}{Tree~\citep{qian2024scaling}} &  \llmname{gemini-1.5-flash} & \textcolor{green(pigment)}{\Checkmark} & \textcolor{darksalmon}{\XSolidBrush} & 81.74 & 94.91 & 71.36 & 77.50 & 73.60 & 79.82 \\

\rowcolor{gray!10} & \llmname{gpt-4o-mini} &  \textcolor{green(pigment)}{\Checkmark} & \textcolor{darksalmon}{\XSolidBrush} & 83.06 & 94.66 & 67.63 & 85.00 & 75.20 & 81.11 \\

\rowcolor{gray!10} \multirow{-2}{*}{Complete Graph~\citep{qian2024scaling}} &  \llmname{gemini-1.5-flash} &  \textcolor{green(pigment)}{\Checkmark}  & \textcolor{darksalmon}{\XSolidBrush} & 81.35 & 94.40 & 68.60 & 83.75 & 74.20 & 80.46\\

\rowcolor{gray!10} & \llmname{gpt-4o-mini} &  \textcolor{green(pigment)}{\Checkmark}  & \textcolor{darksalmon}{\XSolidBrush} & 81.04 & 94.66 & 64.68 & 84.38 & 73.60 & 79.67 \\

\rowcolor{gray!10} \multirow{-2}{*}{LLM-Debate~\citep{arXiv2023_MultiAgent-Debate}} & \llmname{gemini-1.5-flash} &  \textcolor{green(pigment)}{\Checkmark}  & \textcolor{darksalmon}{\XSolidBrush} & 80.40 & 93.98 & 72.45 & 79.38  & 73.40 & 79.92 \\

\rowcolor{gray!10}  & \llmname{gpt-4o-mini} &  \textcolor{green(pigment)}{\Checkmark}  & \textcolor{darksalmon}{\XSolidBrush}  & 82.80 & 94.66 & 68.85 & 86.28 & 75.40 & 81.60\\

\rowcolor{gray!10} \multirow{-2}{*}{GPTSwarm~\citep{zhuge2024gptswarm}} &  \llmname{gemini-1.5-flash} &  \textcolor{green(pigment)}{\Checkmark}  & \textcolor{darksalmon}{\XSolidBrush} & \underline{83.22} & 93.98 & 73.35 & 82.36 & 74.80 & 81.54\\

\rowcolor{gray!10} & \llmname{gpt-4o-mini} &  \textcolor{green(pigment)}{\Checkmark}  & \textcolor{darksalmon}{\XSolidBrush} & 83.02 & \underline{94.89} & 68.45 & 86.80 & 75.40 & 81.71\\

\rowcolor{gray!10} \multirow{-2}{*}{Agentprune~\citep{zhang2024cut}} &  \llmname{gemini-1.5-flash} &  \textcolor{green(pigment)}{\Checkmark}  & \textcolor{darksalmon}{\XSolidBrush} & 83.10 & 93.88 & 73.54 & 82.55 & 75.80 & 81.77\\

\rowcolor{gray!10}  & \llmname{gpt-4o-mini} &  \textcolor{green(pigment)}{\Checkmark}  & \textcolor{darksalmon}{\XSolidBrush}  & 83.10 & 92.30 & 73.35 & \underline{90.06} & \underline{82.20} & \underline{84.20}\\

\rowcolor{gray!10} \multirow{-2}{*}{\cellcolor{gray!10} AFlow~\citep{zhang_aflow_2024}} & \llmname{gemini-1.5-flash} &  \textcolor{green(pigment)}{\Checkmark}  & \textcolor{darksalmon}{\XSolidBrush} & 82.35 & 94.91 & 72.70 & 85.69 & 76.00 & 82.33\\
\hline

PromptLLM~\citep{feng2024graphroutergraphbasedrouterllm} & \llmname{LLM Pool} & \textcolor{darksalmon}{\XSolidBrush} &  \textcolor{green(pigment)}{\Checkmark} & 78.43 & 93.92 & 73.03 & 86.33 & 73.60 & 81.06\\ 

RouteLLM~\citep{ong2024routellmlearningroutellms} & \llmname{LLM Pool} & \textcolor{darksalmon}{\XSolidBrush}  &  \textcolor{green(pigment)}{\Checkmark} & 81.04 & 93.42 & 71.29 & 83.85 & 72.60 & 80.44\\

FragalGPT~\citep{chen2023frugalgptuselargelanguage} & \llmname{LLM Pool} & \textcolor{darksalmon}{\XSolidBrush}  & \textcolor{green(pigment)}{\Checkmark} & 76.24 & 90.76 & 67.05 & 87.31 & 74.40 & 79.15\\

RouterDC~\citep{chen2024routerdc} & \llmname{LLM Pool} & \textcolor{darksalmon}{\XSolidBrush}  & \textcolor{green(pigment)}{\Checkmark} & 82.01 & 93.68 & 73.46 & 87.75 & 75.20 & 82.42\\
\hline

\rowcolor{gray!10} \ourmethod (\textbf{Ours})& \llmname{LLM Pool} & \textcolor{green(pigment)}{\Checkmark}  & \textcolor{green(pigment)}{\Checkmark} & \textbf{84.25} & \textbf{{95.45}} & \textbf{75.42} & \textbf{90.62} & \textbf{84.00} & \textbf{85.93}\\
\Xhline{1.2pt}
\end{tabular}
}
\vspace{-0.3em}
\caption{\small Performance comparison with vanilla, single agent, multi-agent, and single-agent routing methods. The best results are highlighted in bold, and the runner-ups are underlined. The LLM pool includes the economical and advanced LLMs mentioned in \Cref{para:llmbackbones}. "Mul." and "Rout." indicate whether the method supports a multi-agent setting and whether it supports the LLM routing, respectively. \textcolor{darksalmon}{\XSolidBrush} and \textcolor{green(pigment)}{\Checkmark} indicate whether these features are supported.}
\vspace{-0.5em}
\label{tab:rq1_performance}
\vspace{-0.5em}
\end{table*}

\subsection{Optimization}
\label{sec:methodoptim}
The optimization objective of \ourmethod is presented as follows:
\begin{equation}\label{eq:objective}
\underset{\theta}{\min} \;\mathbb{E}_{\substack{(\mathcal{Q},a)\sim \mathcal{D}, \\\mathcal{S}\sim\mathbb{F}_\theta}} \big[-p(a | \mathcal{Q}) + \lambda\cdot C(\mathcal{S};\mathcal{Q})\big]
\end{equation}
where \(C(\cdot)\) represents the cost evaluation of multi-agent systems, and \(\lambda\) is the trade-off parameter. The term \(p(a | \mathcal{Q})\) in \Cref{eq:objective} corresponds to \Cref{eq:core}, which was computed in the previous sections. Through this optimization objective, we balance effectiveness and efficiency by maximizing the probability of generating correct solutions while minimizing token expenditure. Then following standard approaches in multi-agent structure design~\citep{zhuge2024gptswarm,zhang2024cut}, we apply policy gradient~\citep{williams1992simple} to approximate and optimize \Cref{eq:objective}.

We summarize the notations in \Cref{app:notation}, with the algorithmic workflow in \Cref{app:workflow}.

\vspace{-0.3em}
\section{Experiments}
\vspace{-0.3em}
\subsection{Experimental Setup}
\paragraph{Dataset and Benchmarks}
We opt for \textbf{MMLU}~\citep{mmlu}, \textbf{GSM8K}~\citep{arXiv2021_Verifier-Math}, \textbf{MATH}~\citep{hendrycksmath2021}, \textbf{HumanEval}~\citep{human-eval}, \textbf{MBPP}~\citep{mbpp}, covering a diverse range of reasoning and problem-solving tasks. For the MATH dataset, we select 519 problems from different levels using stratified sampling.

\vspace{-0.4em}
\paragraph{Baselines}
We compare our method with (1) single-agent approaches, including \textbf{COT}~\citep{cot}, \textbf{ComplexCoT}~\citep{fu2022complexity}, \textbf{Self-Consistency}~\citep{ICLR2023_Self-Consistency}; (2) fixed multi-agent topologies including \textbf{Chain}, \textbf{Tree}, and \textbf{Complete Graph}  (formally defined in ~\citep{qian2024scaling}), \textbf{LLM-Debate} ~\citep{chateval}; (3) dynamic multi-agent systems like \textbf{GPTSwarm}~\citep{zhuge2024gptswarm}, \textbf{AgentPrune}~\citep{zhang2024cut} and \textbf{AFlow}~\citep{zhang_aflow_2024}; (4) single LLM routers including \textbf{PromptLLM} introduced in \cite{feng2024graphroutergraphbasedrouterllm}, \textbf{RouteLLM}~\citep{ong2024routellmlearningroutellms}, \textbf{FrugalGPT}~\citep{chen2023frugalgptuselargelanguage} and \textbf{RouterDC}~\citep{chen2024routerdc}.

\vspace{-0.4em}
\paragraph{LLM Backbones} \label{para:llmbackbones}
We select LLMs with varying sizes and capacities, including \llmname{gpt-4o-mini-0718}~\citep{OpenAI-gpt4o}, \llmname{claude-3.5-haiku}~\citep{anthropicdd2024claude}, \llmname{gemini-1.5-flash}~\citep{geminiteam2024gemini15unlockingmultimodal} and \llmname{llama-3.1-70b}~\citep{dubey2024llama} as the llm pool. \llmname{Deepseek-v3}~\citep{deepseekai2024deepseekv3technicalreport} is used to validate \ourmethod's inductive capabilities. The temperature is always set as $1$.

\vspace{-0.4em}
\paragraph{Implementation Details}
We set the learning rate \( \alpha = 0.01 \), the temperature \( \tau = 1 \), the cost penalty \( \lambda \in \{5,15,25\} \) and the num of iteration \(K \in \{5,10\}\), the agent's maximum amount \( \gamma = 6 \). In the MAS baselines, the number of agents equals $\gamma$. We use all the LLMs mentioned in the LLM Backbones as the \textit{candidate LLM pool} for the routing method. The collaboration modes repository includes CoT, Reflection, self-consistency, LLM debate, and Macnet (Chain \& Complete graph). The role pool comprises 26 roles with diverse capabilities, such as programmers using compilers and researchers with access to Wikipedia. The details of the candidate pools can be found in \Cref{app:profile}. 

\subsection{Performance \& Cost Analysis}
In this section, we compare \ourmethod with \underline{twenty baselines} across five benchmarks. We verify that \ourmethod is:

\vspace{-0.4em}
\paragraph{High-performing.} The experimental results in \Cref{tab:rq1_performance} demonstrates that \ourmethod excels at constructing an effective multi-agent system. Specifically, \ourmethod achieves the best performance across all of the five datasets, outperforming RouterDC, the SOTA LLM routing method, by $3.51\%$ on average. On the MBPP dataset, \ourmethod outperforms AgentPrune and AFlow by $8.20\%$ and $1.80\%$ at pass@1, respectively.

\vspace{-0.4em}
\paragraph{Token-economical.}
\begin{figure}[!t]
\centering
\includegraphics[width=1.0\linewidth]{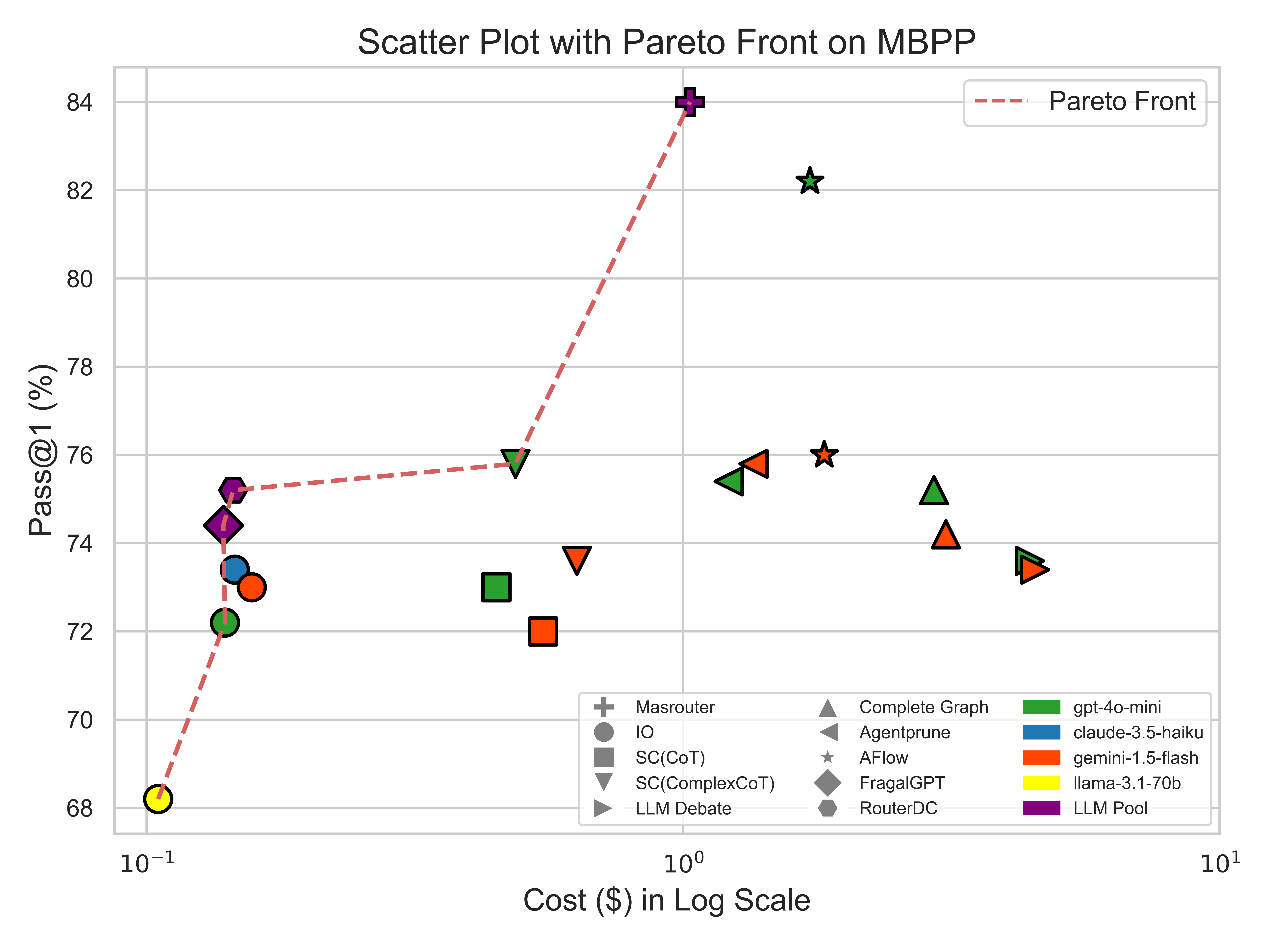}
\vspace{-2em}
\caption{The comparison of the performance and inference cost on the MBPP dataset. Different shapes of the scatter points represent various types of baselines, while the different colors of the points indicate the use of different LLM backbones.}
\vspace{-0.6em}
\label{fig:rq2-cost}
\end{figure}
As shown in \Cref{fig:rq2-cost}, \ourmethod achieves the best performance on the Pareto front of cost-effectiveness on the MBPP dataset. Compared to AFlow, \ourmethod not only achieves a $1.8\%\sim 8.0\%$ improvement in performance but also reduces the inference overhead by $40.22\%\sim 43.78\%$. 

\vspace{-0.4em}
\paragraph{Training resource-saving.} As shown in \Cref{tab:trainingcost}, compared to trainable MAS pipelines like GPTSwarm and AFlow, we achieve savings of $69.57\%$ and $83.51\%$ on the MMLU dataset, respectively. This is because our method does not require exhaustive traversal and validation of each agentic structure. We present the detailed cost-performance data in \Cref{app:cost}.

\vspace{-0.3em}
\subsection{Plug-in to Existing MAS}
\vspace{-0.3em}

\begin{table}[t!]
  \centering
  \vspace{-0.5em}
  \resizebox{\linewidth}{!}{
  \begin{tabular}{l|cccc} 
    \Xhline{1.2pt}
    \rowcolor{CadetBlue!20} 
    \textbf{Dataset} & \textbf{Method} & \textbf{LLM} & \textbf{Performance} & \textbf{Cost}\\
    \Xhline{1pt}
    \multirow{3}{*}{MMLU} & \multirow{2}{*}{MAD} & \llmname{gpt} & $81.50$ & $\$25.56$ \\
    & & \llmname{gemini} & $80.94$ & $\$27.02$ \\
    \cline{2-5}
    & \multicolumn{2}{c}{+\ourmethod} & $82.20({\color{RedOrange} \uparrow 0.70})$ &  $\$19.39$\\
    \hline
   \rowcolor{gray!10}& & \llmname{gpt} & $86.05$ & $\$1.248$ \\
   \rowcolor{gray!10}& \multirow{-2}{*}{MAD} & \llmname{gemini} & $82.95$ & $\$1.526$ \\
   \cline{2-5}
   \rowcolor{gray!10} \multirow{-3}{*}{HumanEval}& \multicolumn{2}{c}{+\ourmethod} & $87.60({\color{RedOrange} \uparrow 1.55})$ &  $\$1.096$\\
    \hline
    \multirow{3}{*}{GSM8K} & \multirow{2}{*}{MAD} & \llmname{gpt} & $94.60$ & $\$5.664$ \\
    & & \llmname{gemini} & $94.40$ & $\$5.492$ \\
    \cline{2-5}
    & \multicolumn{2}{c}{+\ourmethod} & $94.91({\color{RedOrange} \uparrow 0.31})$ &  $\$4.702$\\
    \hline
   \rowcolor{gray!10}& & \llmname{gpt} & $82.98$ & $\$7.812$ \\
   \rowcolor{gray!10}& \multirow{-2}{*}{MacNet} & \llmname{gemini} & $81.74$ & $\$8.482$ \\
   \cline{2-5}
   \rowcolor{gray!10} \multirow{-3}{*}{MMLU}& \multicolumn{2}{c}{+\ourmethod} & $83.40({\color{RedOrange} \uparrow 0.36})$ &  $\$5.892$\\
   \hline
    \multirow{3}{*}{HumanEval} & \multirow{2}{*}{MacNet} & \llmname{gpt} & $86.82$ & $\$0.488$ \\
    & & \llmname{gemini} & $83.72$ & $\$0.568$ \\
    \cline{2-5}
    & \multicolumn{2}{c}{+\ourmethod} & $88.37({\color{RedOrange} \uparrow 1.55})$ &  $\$0.404$\\
    \hline
   \rowcolor{gray!10}& & \llmname{gpt} & $94.69$ & $\$2.142$ \\
   \rowcolor{gray!10}& \multirow{-2}{*}{MacNet} & \llmname{gemini} & $94.31$ & $\$2.016$ \\
   \cline{2-5}
   \rowcolor{gray!10} \multirow{-3}{*}{GSM8K}& \multicolumn{2}{c}{+\ourmethod} & $94.89({\color{RedOrange} \uparrow 0.20})$ &  $\$1.774$\\
    \Xhline{1.2pt}
  \end{tabular}
  }
  \vspace{-0.3em}
  \caption{Comparison of performance and cost before and after integrating with \ourmethod. \llmname{gpt} and \llmname{gemini} are abbreviations for \llmname{gpt-4o-mini} and \llmname{gemini-1.5-flash}, respectively. The MacNet method uses the optimal structure reported in the paper.} 
  \vspace{-0.5em}
  \label{tab:plugin}
\end{table}

Considering that contemporary MAS is often LLM-homogeneous, \textit{i.e.}, relying exclusively on a single powerful model like \llmname{gpt-4o}, \ourmethod can serve as a plug-and-play solution, seamlessly assigning an optimal LLM backbone to each agent within them, resulting in significantly less inference cost and comparable performance. In \Cref{tab:plugin}, we combine \ourmethod with the well-established MAS methods MAD~\citep{arXiv2023_MultiAgent-Debate} and MacNet. \ourmethod improves the performance of MAD by $1.55\%$ at pass@1 on the Humaneval dataset while reducing cost by $17.21\% \sim 28.17\%$. On larger datasets, the significant reduction in overhead is even more notable; when integrated with MAD, \ourmethod saves the inference cost by $6.17 \sim 7.63\$$ on MMLU. Overall, \ourmethod can serve as a plugin to support economical multi-agent development. 

\subsection{Inductive Ability Analysis}
\begin{figure}[t]
\centering
\includegraphics[width=0.85\linewidth]{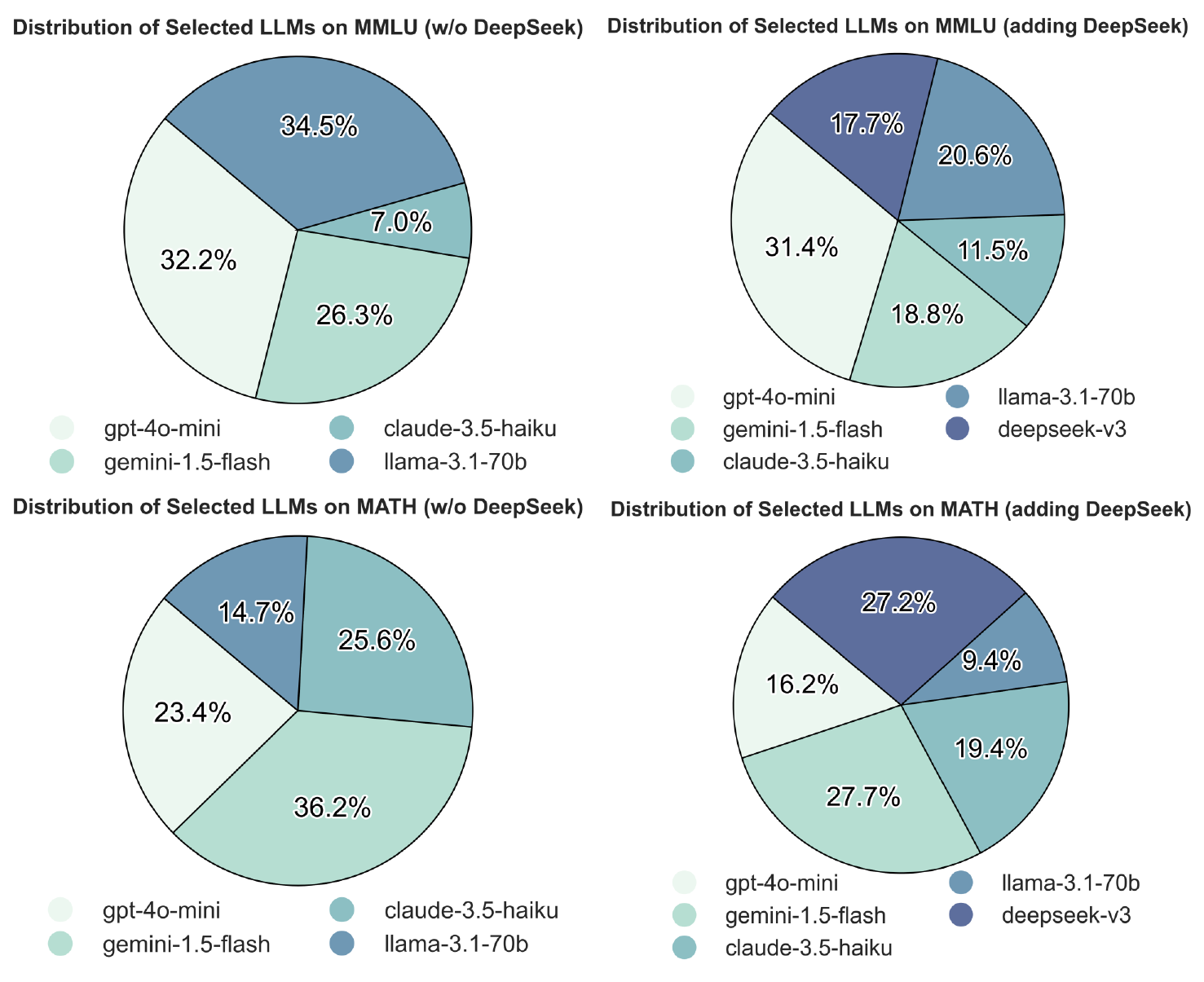}
\vspace{-1em}
\caption{The selected LLM distribution of \ourmethod on MMLU and MATH benchmark.}
\vspace{-1em}
\label{fig:inductive}
\end{figure}

In this section, we validate that \ourmethod can easily generalize to unseen LLMs without intensive re-training resources. \Cref{fig:inductive} illustrates the distribution of LLMs selected by \ourmethod before and after the addition of \llmname{Deepseek-v3} on MMLU and MATH, with the new model being chosen $12.17\%$ and $27.19\%$ the time, respectively. By intelligently selecting the new, stronger model, \ourmethod improved the accuracy on MMLU from $84.25\%$ to $85.40\%$ and also improved the accuracy on HumanEval from $90.62\%$ to $91.41\%$. 

\begin{table}[!htpb]
\vspace{-1em}
\vspace{0.1em}
\centering
\resizebox{\columnwidth}{!}{
\begin{tabular}{c|cc|cc}
\toprule
\makecell{Dataset} &\multicolumn{2}{c|}{GSM8K} & \multicolumn{2}{c}{MATH}\\
\midrule
\makecell{Metric}  &  \makecell{Accuracy \\(\%)} & \makecell{Cost\\ ($\$$)}   &  \makecell{Accuracy \\(\%)} & \makecell{Cost\\ ($\$$)}  \\
\midrule
\makecell{Vanilla \ourmethod} & $95.45$ & $1.59$ & $75.42$ & $3.58$ \\
\midrule
\ourmethod \textit{w/o} \(\mathbb{F}_{\theta_t}\)  & $93.84$ & $2.38$ & $72.77$ & $4.48$ \\
\ourmethod \textit{w/o} \(\mathbb{F}_{\theta_r}\) & $94.70$ & $1.67$ &  $73.01$ & $3.63$   \\
\ourmethod \textit{w/o} \(\mathbb{F}_{\theta_m}\) & $93.36$ & $1.98$ & $71.08$ & $4.16$\\
\ourmethod \textit{w/o} $C(\cdot)$ & $95.63$ & $2.45$ & $75.18$ & $5.07$\\
\bottomrule
\end{tabular}}
\caption{Ablation study of \ourmethod.}\label{tab:ablation}
\vspace{-0.8em}
\end{table}

\subsection{Framework Analysis}
\vspace{-0.3em}
\paragraph{Ablation Study} We conduct an ablation study on the four key modules in \ourmethod: \textbf{(1) \textit{w/o} \(\mathbb{F}_{\theta_t}\)}, which replaces the collaboration determination \(\mathbb{F}_{\theta_t}\) with random selection, \textbf{(2) \textit{w/o} \(\mathbb{F}_{\theta_r}\)}, which replaces the role allocator \(\mathbb{F}_{\theta_r}\) with random selection, \textbf{(3) \textit{w/o} \(\mathbb{F}_{\theta_m}\)}, which replaces the LLM router \(\mathbb{F}_{\theta_r}\) with random selection, and \textbf{(4) \textit{w/o} \(C(\cdot)\)}, which remove the cost evaluation in \Cref{eq:objective}. As shown in \Cref{tab:ablation}, removing the $\mathbb{F}_{\theta_m}$ results in the largest performance decline by $2.09\%$ and $4.34\%$. This is because the performance of the base models on the dataset varies significantly, making selecting the appropriate LLM to solve the problem crucial. Removing \(C(\cdot)\) does not significantly impact the performance, but it disrupts the adaptive capability of \ourmethod to query difficulty, leading to an increase in overhead by $54.09\%$ and $41.62\%$. 

\paragraph{Sensitivity Analysis}
\begin{figure}[!t]
\centering
\includegraphics[width=1.0\linewidth]{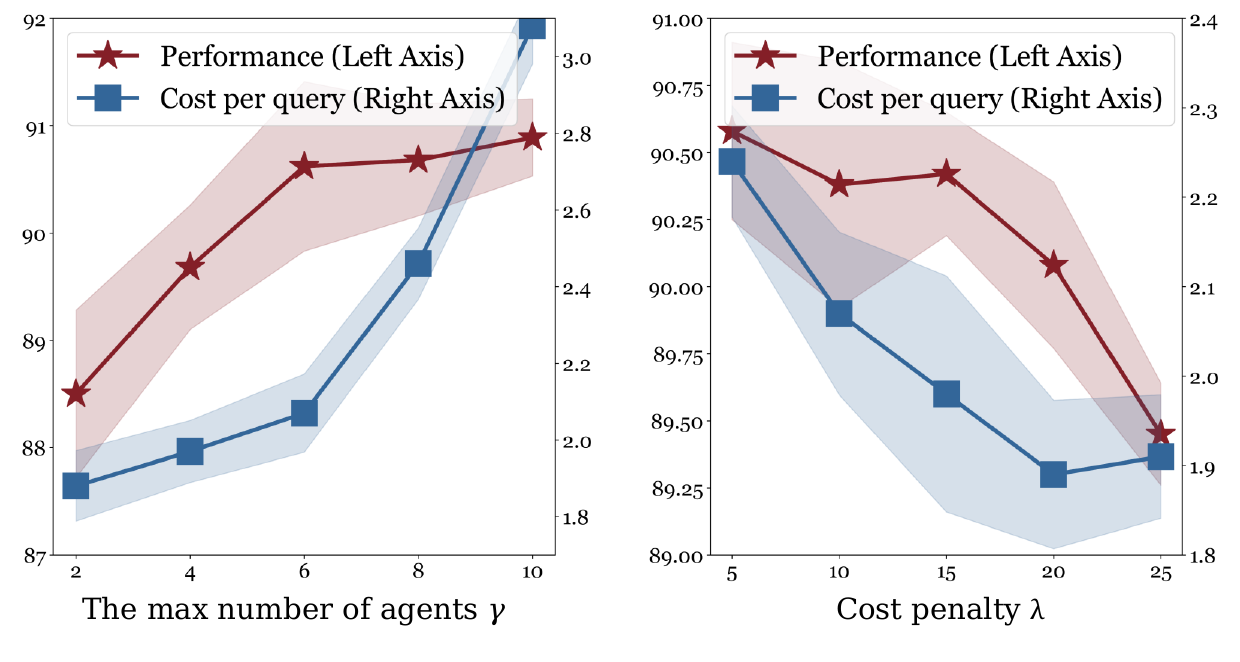}
\vspace{-2em}
\caption{Sensitivity analysis of \ourmethod on HumanEval. The unit of cost per query (right) and performance (left) is $10^{-3}\cdot\$$ and $pass@1~(\%)$, respectively.
}
\vspace{-0.9em}
\label{fig:sensi}
\end{figure}
We analyze the sensitivity of \ourmethod to two core parameters: the maximum number of the agents $\gamma$, the cost penalty coefficient $\lambda$ in \Cref{eq:objective}. The results are presented in \Cref{fig:sensi}. 
\textbf{For the parameter $\gamma$}, we observe a significant performance improvement as $\gamma$ increases from 2 to 6 ($88.50\% \rightarrow 90.62\%$). However, further increases from 6 to 10 yield only marginal performance gains while incurring the $1.5\times$ per-query inference costs. Considering both performance and cost, we select $\gamma = 6$. \textbf{For the parameter $\lambda$}, as $\lambda$ increases from 5 to 25, we find that larger values lead \ourmethod to favor more cost-efficient solutions, reducing the overhead by 17.78\%, albeit with a slight performance degradation of approximately $1.3\%$. We balance effectiveness and cost by dynamically adjusting this value.

\subsection{Case Study}
\vspace{-0.3em}
We present a detailed case study and visualization of the routing process of \ourmethod across five benchmarks. We sincerely refer the readers to \Cref{app:case} for details.

\section{Conclusion}
\vspace{-0.5em}
In this paper, we \textit{for the first time} introduce \textbf{Multi-Agent System Routing (MASR)}, which aims to intelligently allocate collaboration patterns, agent roles, and LLMs for each query, thereby constructing a customized MAS. Based on this concept, we present \ourmethod, the first high-performing, economical, and inductive MASR solution. \ourmethod progressively builds a list of mutually adaptive roles, selects an LLM proficient in the task domain, and ultimately achieves a balance between effectiveness and efficiency. We believe \ourmethod paves the way for the automation, economization, and scalability of MAS, contributing to the development of large-scale collective intelligence. 

\bibliography{arxiv}

\appendix

\section{Notations}\label{app:notation}
We conclude the commonly used notations in \Cref{tab:notation} for reference.

\begin{table*}[!h]\footnotesize
  \centering
   \setlength{\tabcolsep}{10pt}
   \renewcommand\arraystretch{1.15}
  \vspace{-2mm}
    \begin{tabular}{cc}
        \Xhline{1.2pt}
    \rowcolor{CadetBlue!20} 
    \textbf{Notation} & \textbf{Definition} \\
     \Xhline{1.pt}
    $\mathbb{S} = \{\mathbb{M}, \mathbb{R}, \mathbb{T}\}$ & Candidate space containing LLM pool $\mathbb{M}$, roles set $\mathbb{R}$, and collaboration modes set $\mathbb{T}$ \\
    $\mathbb{M}$ & Pool of available LLM backbones \\
    $\mathbb{R}$ & Set of predefined agent roles (e.g., Analyst, Developer, Tester) \\
    $\mathbb{T}$ & Set of collaboration modes (e.g., Chain, Tree, Debate) \\
    $k$ & Number of agents in the multi-agent system \\
    $\mathcal{S} = \{\{\mathcal{M}_i\}_{i=1}^k, \{\mathcal{R}_i\}_{i=1}^k, \mathcal{T}\}$ & Multi-agent system with LLMs $\mathcal{M}_i$, roles $\mathcal{R}_i$, and collaboration mode $\mathcal{T}$ \\
    $\mathcal{M}_i$ & Selected LLM backbone for the $i$-th agent \\
    $\mathcal{R}_i$ & Selected role for the $i$-th agent \\
    $\mathcal{Q}$ & Input query to the multi-agent system \\
    $a$ & Oracle answer corresponding to the query $\mathcal{Q}$ \\
    $f: \mathcal{M} \times \mathcal{R} \times \mathcal{T} \rightarrow \mathcal{S}$ & MASR mapping function assigning components to queries \\
    $\pi(\mathcal{S})$ & Probability of selecting system $\mathcal{S}$ given query $Q$ \\
    $U(\mathcal{S}; \mathcal{Q}, a)$ & Utility function measuring MAS performance \\
    $C(\mathcal{S}; \mathcal{Q})$ & Cost function quantifying token expenditure \\
    $\lambda$ & Trade-off parameter between utility and cost \\
    $\mathbb{F}_\theta = \mathbb{F}_{\theta_t} \circ \mathbb{F}_{\theta_r} \circ \mathbb{F}_{\theta_m}$ & Controller network for collaboration, role allocation, and LLM routing \\
    $\mathbf{H}$ & Latent variable capturing query-collaboration semantics \\
    $\tilde{\mathbf{H}}_\tau$ & Refined representation of the candidate collaboration mode $\mathcal{T}$ \\
    $\tau$ & Temperature parameter in probability decoding \\
    $\gamma$ & Hyperparameter for maximum number of agents \\
    $p(a|\mathcal{Q})$ & Conditional likelihood of generating answer $a$ via MAS \\
    $\Gamma(z)$ & Gamma function approximating non-integer factorials \\
    $\delta(\mathbf{H})$ & Complexity mapping function determining the number of agents \\
    $f_\psi(\mathcal{\cdot})$ & Encoder extracting semantic information from the query $\mathcal{Q}$ \\
    $g_\phi(\cdot)$ & Fusion module producing refined representations \\
    \Xhline{1.2pt}
    \end{tabular}%
   \caption{The notations that are commonly used throughout the manuscript.}
     \label{tab:notation}%
\end{table*}%

\section{Algorithm Workflow}\label{app:workflow}
We conclude the overall algorithm workflow of \ourmethod in Algorithm \ref{algo:masrouter}.

\begin{algorithm*}
\caption{Workflow of \ourmethod}
\label{algo:masrouter}
\Input{Benchmark $\mathcal{D}$, encoder $f_{\psi}(\cdot)$, fusion module $g_\phi(\cdot)$, learning rate $\alpha$, search space $\mathbb{S} = \{\mathbb{M},\mathbb{R},\mathbb{T}\}$}
\For{$\text{query}$ $\mathcal{Q} \in \mathcal{D}$}{
    \For{$\text{iteration}$ $t \in \{1,2,\cdots,K\}$}{
        \textcolor{blue}{\tcc{Collaboration Mode Determination}}
        
         Sample latent vector $\mathbf{H} \sim \mathcal{N}(\mu_t(\mathcal{Q}), \text{diag}(\sigma_t^2(\mathcal{Q})))$ $\quad$ \textcolor{gray}{ $\triangleright$ {Eq.(7)}}
         
         Compute collaboration mode probability: $p(\mathcal{T}|\mathbf{H}) \propto \exp(f_\psi(\mathcal{Q})^\top \tilde{\mathbf{H}}_\tau/\tau)$
         
       Determine agent count: $k = \lceil \delta(\mathbf{H}) \cdot \gamma \rceil$ $\quad$ \textcolor{gray}{$\triangleright$ Dynamic scaling}
        
        \textcolor{blue}{\tcc{Agent Role Allocation}}
        \For{$\ell = 1$ \textbf{to} $k$}{
             Compute role probability: $\pi_{r\ell} \propto \exp(\mathbf{H}_{\mathcal{R}_{\ell-1}}^\top \mathbf{H}_{\mathcal{R}_\ell}/\tau)$
            
            Select role $\mathcal{R}_\ell$ via cascaded inference $\quad$ \textcolor{gray}{// Eq.(9)}
        }
        
        \textcolor{blue}{\tcc{Agent LLM Routing}}
        
        Aggregate context: $\mathbf{H}_{\mathcal{M}} = \text{FFN}(\mathbf{H} \oplus \mathbf{H}_{\mathcal{T}} \oplus \sum \mathbf{H}_{\mathcal{R}_i})$
        
        \For{each agent $i \in \{1,...,k\}$}{
             Compute LLM compatibility: $\pi_m(\mathcal{M}_i) \propto \exp(\mathbf{H}_{\mathcal{M}}^\top \mathbf{H}_{\mathcal{M}_i}/\tau)$ $\quad$ \textcolor{gray}{// Eq.(11)}
            
            Assign LLM $\mathcal{M}_i$ with multinomial sampling
        }
        
        \textcolor{blue}{\tcc{Optimization}}
        
        Compute reward $R = U(\mathcal{S};\mathcal{Q},a) - \lambda \cdot C(\mathcal{S};\mathcal{Q})$ $\quad$ \textcolor{gray}{// Eq.(3)}
        
        Update $\theta$ via policy gradient: $\theta \leftarrow \theta - \alpha \nabla_\theta \mathbb{E}[-R]$ $\quad$ \textcolor{gray}{// Section 4.4}
    }
}
\end{algorithm*}

\section{Case Study}\label{app:case}
As shown in \Cref{tab:mmlucase,tab:gsmcase,tab:mathcase,tab:humanevalcase,tab:mbppcase}, we visualize the customized MAS designed by \ourmethod for varying query difficulties on the five benchmarks.

\begin{table*}[ht]
\resizebox{\linewidth}{!}{
\begin{tabular}{m{0.55\linewidth}|m{0.45\linewidth}}
\hline
\multicolumn{1}{c|}{\textbf{Query}} & \multicolumn{1}{c}{\textbf{MasRouter Workflow}} \\
\hline
     Bentham defines the fecundity of a pleasure or pain as:
Option A: its chance of occurring.
Option B: the degree to which it is felt.
Option C: its chance of being followed by sensations of the same kind.
Option D: how long it lasts. & \adjustbox{valign=c}{\includegraphics[width=\linewidth]{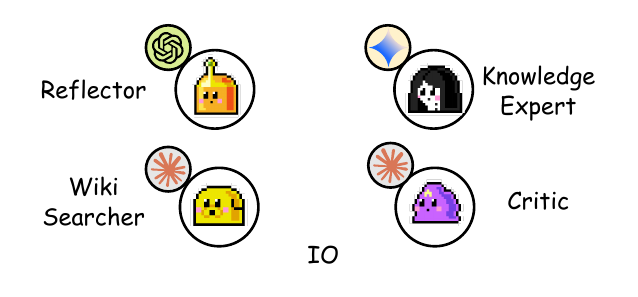}} \\
\hline
This jurisdiction has the following bribery statute in effect:"Any person who offers or gives a thing of value to a government officeholder in exchange for official action is guilty of bribery. "A real estate developer owned a large parcel of land in the suburbs. Although the developer wanted to build an office building on the property, the land was zoned residential. Due to the residential zoning, the developer could not pursue his planned development unless he received a variance from the building commission. The developer held a meeting with a member of the building commission to solicit his approval in securing a zoning variance. To do so, the developer gave the commission member \$10,000 in exchange for his support in approving the zoning variance. Thereupon, the commission member voted to approve the variance, thus making it possible for the developer to commence construction of the office building. The developer was subsequently prosecuted for conspiracy to commit bribery. During the course of the trial, the commission member testified that he faked the agreement with the developer and would have approved the zoning variance regardless of whether the developer gave him any money. Furthermore, in his defense, the developer presented evidence that the other six members of the building commission voted affirmatively to approve the variance. If the jury believed that the commission member would have approved the variance even had he not received the \$10,000, the developer should be found 
Option A: guilty, because the commission member's agreement to accept the \$10,000 was sufficient to form a conspiratorial objective.
Option B: guilty, because he gave the commission member the \$10,000 in exchange for his approval of the zoning variance.
Option C: not guilty, because the commission member did not receive a thing of value, since he would have approved the variance regardless of receiving any payment from the developer.
Option D: not guilty, because there was no true agreement between the parties. & 
\adjustbox{valign=c}{\includegraphics[width=\linewidth]{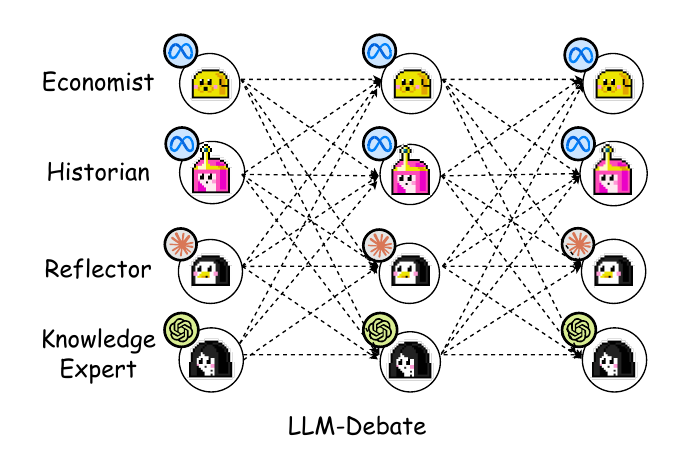}}

\\
\hline
\end{tabular}
}
\caption{MMLU dataset}
\label{tab:mmlucase}
\end{table*}

\begin{table*}[ht]
\resizebox{\linewidth}{!}{
\begin{tabular}{m{0.55\linewidth}|m{0.45\linewidth}}
\hline
\multicolumn{1}{c|}{\textbf{Query}} & \multicolumn{1}{c}{\textbf{MasRouter Workflow}} \\
\hline
    Grandma Jones baked 5 apple pies for the fireman's luncheon.  She cut each pie into 8 pieces and set the five pies out on the buffet table for the guests to serve themselves.  At the end of the evening, after the guests had taken and eaten their pieces of pie, there were 14 pieces of pie remaining.  How many pieces were taken by the guests? & \includegraphics[width=\linewidth]{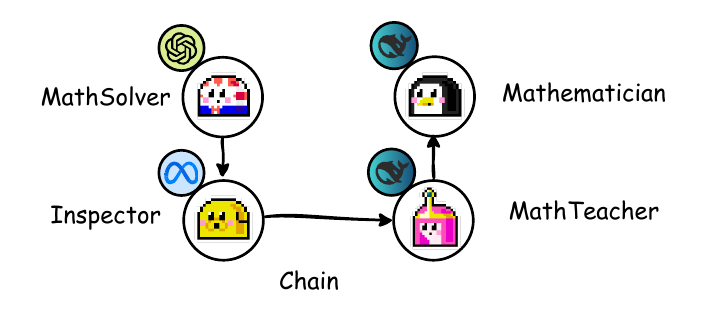} \\
\hline
The combined age of Peter, Paul and Jean is 100 years old. Find the age of Peter knowing that Paul is 10 years older than John and that Peter’s age is equal to the sum of Paul and John's age.& \includegraphics[width=\linewidth]{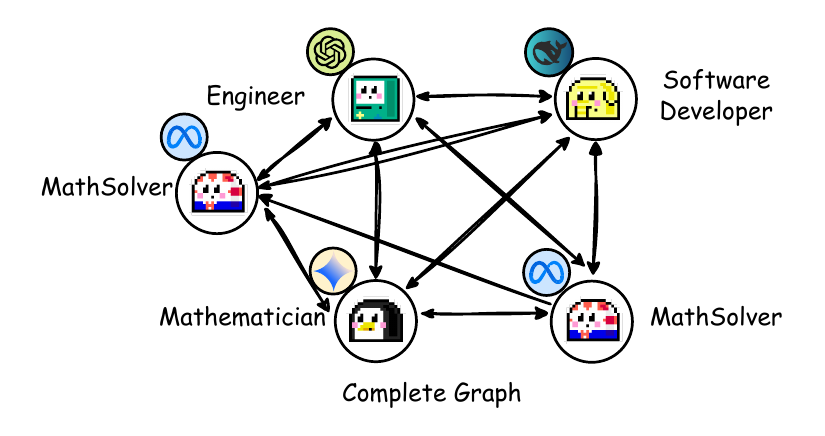} \\
\hline
\end{tabular}

}
\caption{GSM8K dataset}
\label{tab:gsmcase}
\end{table*}

\begin{table*}
\resizebox{\linewidth}{!}{
\begin{tabular}{m{0.55\linewidth}|m{0.45\linewidth}}
\hline
\multicolumn{1}{c|}{\textbf{Query}} & \multicolumn{1}{c}{\textbf{MasRouter Workflow}} \\
\hline
    What is the value of $(4 \times 12)-(4+12)$? & \includegraphics[width=\linewidth]{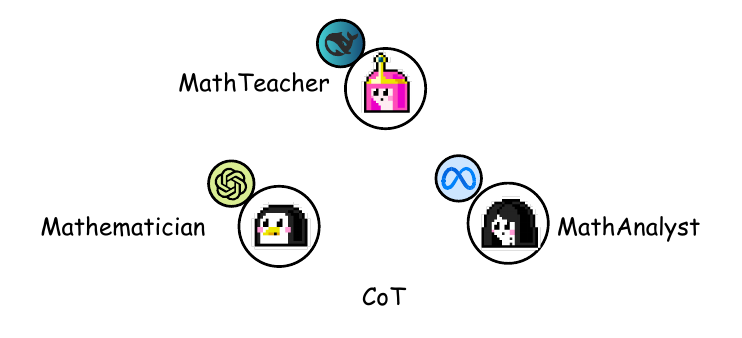} \\
\hline
In the diagram, $K$, $O$ and $M$ are the centers of the three semi-circles. Also, $OC = 32$ and $CB = 36$. [asy]
pair A, K, O, C, M, B, X, Y, Z;
O=(0,0);
C=(32,0);
M=(50,0);
B=(68,0);
A=(-68,0);
K=(A+C)/2;
X=(0,68);
Y=(-18,50);
Z=(50,18);
path nom, bigc, middlec, smallc;
nom=A--B--(100,100)--(-100,100)--cycle;
bigc=A..X..B--cycle;
middlec=A..Y..C--cycle;
smallc=C..Z..B--cycle;
fill(bigc, gray(.5));
fill(middlec, white);
fill(smallc, white);
draw(smallc);
draw(middlec);
draw(bigc);
draw(A--B);
label("A", A, S);
label("K", K, S);
label("O", O, S);
label("M", M, S);
label("C", C, S);
label("B", B, S);
dot(K);
dot(O);
dot(M);
[/asy]
What is the length of \$AC\$?& \includegraphics[width=\linewidth]{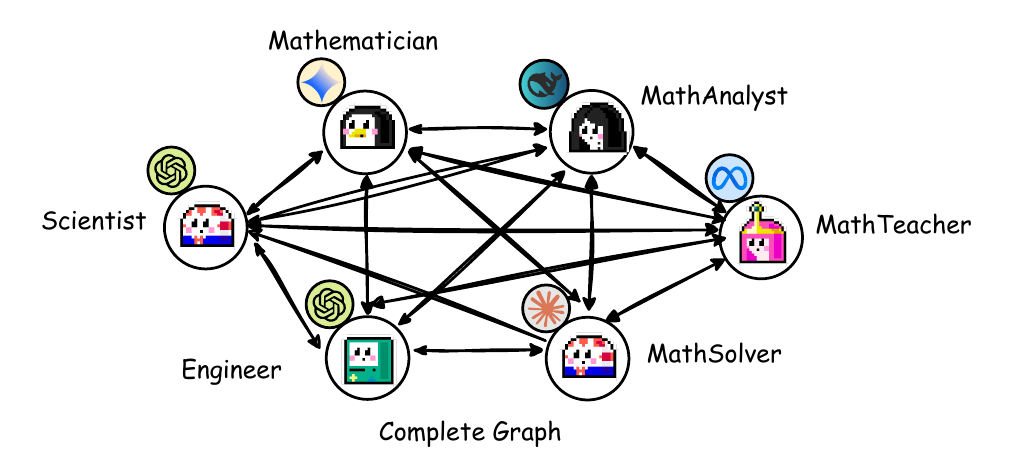} \\
\hline
\end{tabular}
}
\caption{MATH dataset}
\label{tab:mathcase}
\end{table*}

\begin{table*}
\lstset{linewidth=0.55\linewidth}
\resizebox{\linewidth}{!}{
\begin{tabular}{m{0.55\linewidth}|m{0.45\linewidth}}
\hline
\multicolumn{1}{c|}{\textbf{Query}} & \multicolumn{1}{c}{\textbf{MasRouter Workflow}} \\
\hline
    \includegraphics[width=\linewidth]{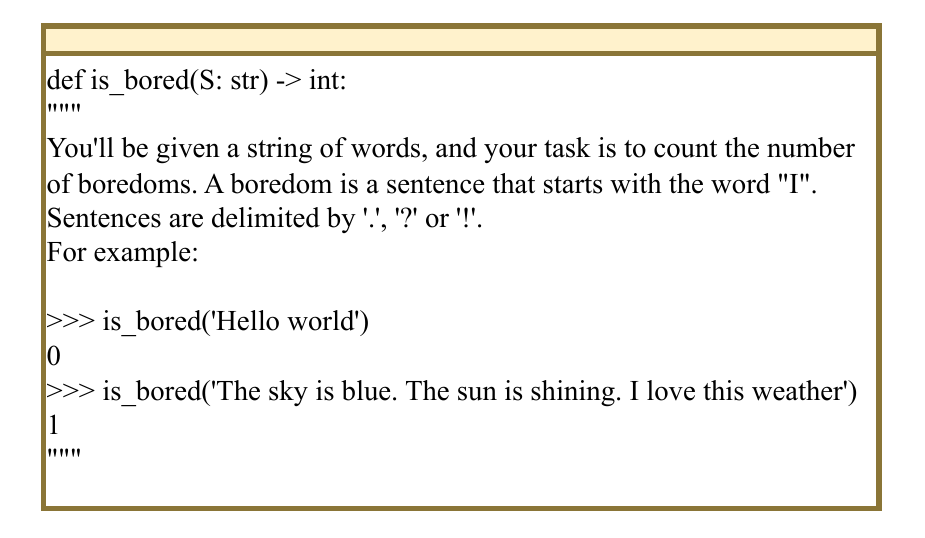}
    & \includegraphics[width=\linewidth]{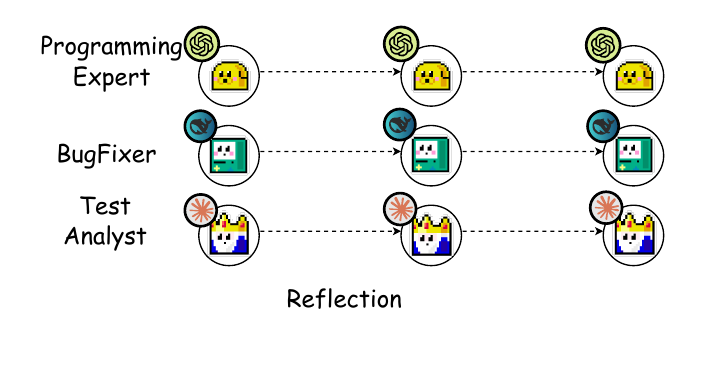} \\
\hline
    \includegraphics[width=\linewidth]{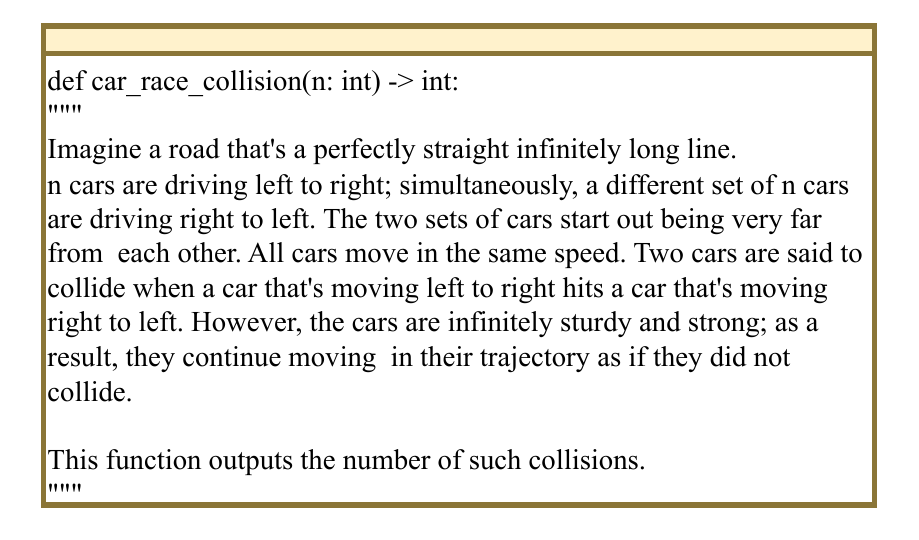}& 
    \includegraphics[width=\linewidth]{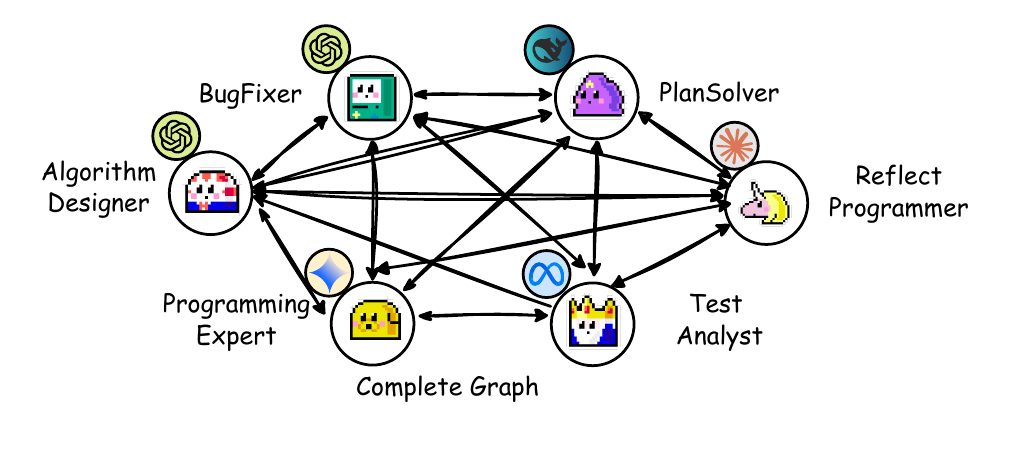} 
    \\
\hline
\end{tabular}
}
\caption{HumanEval dataset}
\label{tab:humanevalcase}
\end{table*}

\begin{table*}
\resizebox{\linewidth}{!}{
\begin{tabular}{m{0.55\linewidth}|m{0.45\linewidth}}
\hline
\multicolumn{1}{c|}{\textbf{Query}} & \multicolumn{1}{c}{\textbf{MasRouter Workflow}} \\
\hline
    \includegraphics[width=\linewidth]{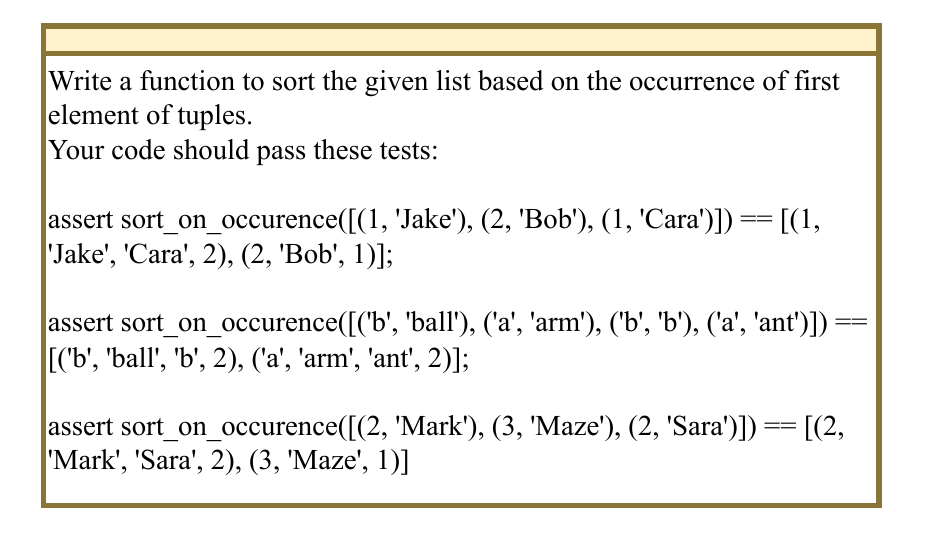}&   
    \includegraphics[width=\linewidth]{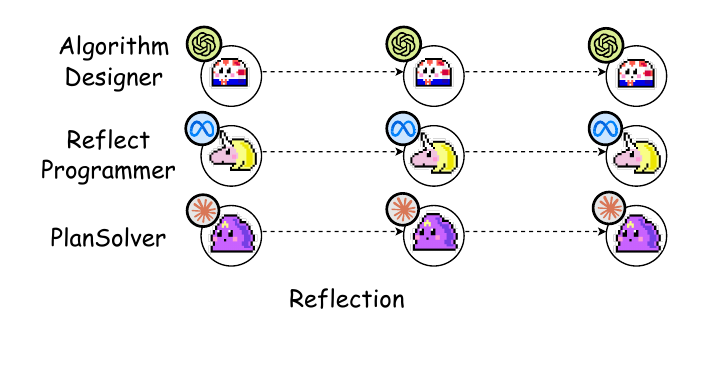} \\
\hline
    \includegraphics[width=\linewidth]{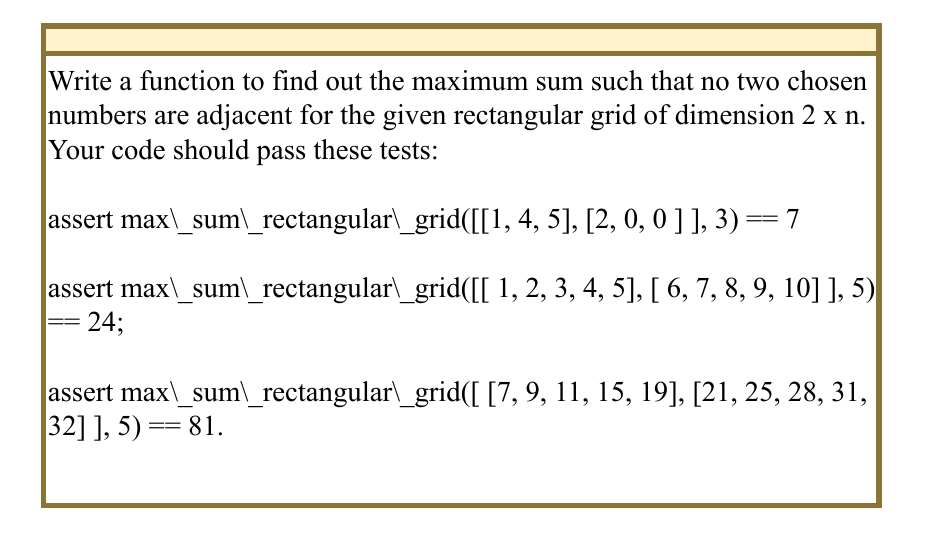}& \includegraphics[width=\linewidth]{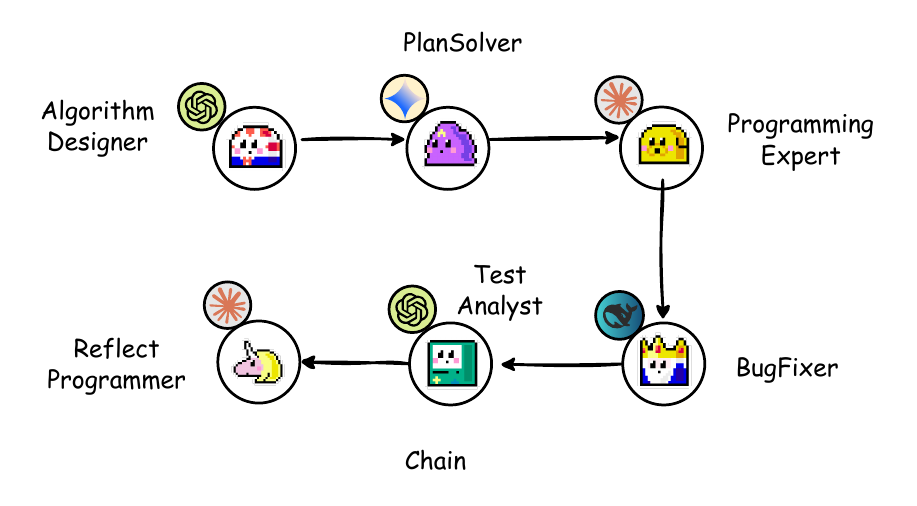} \\
\hline
\end{tabular}
}
\caption{MBPP dataset}
\label{tab:mbppcase}
\end{table*}

\section{Detailed Cost-Performance Data}\label{app:cost}
\subsection{Inference Cost}
\begin{figure}[!ht]
\centering
\includegraphics[width=1.0\linewidth]{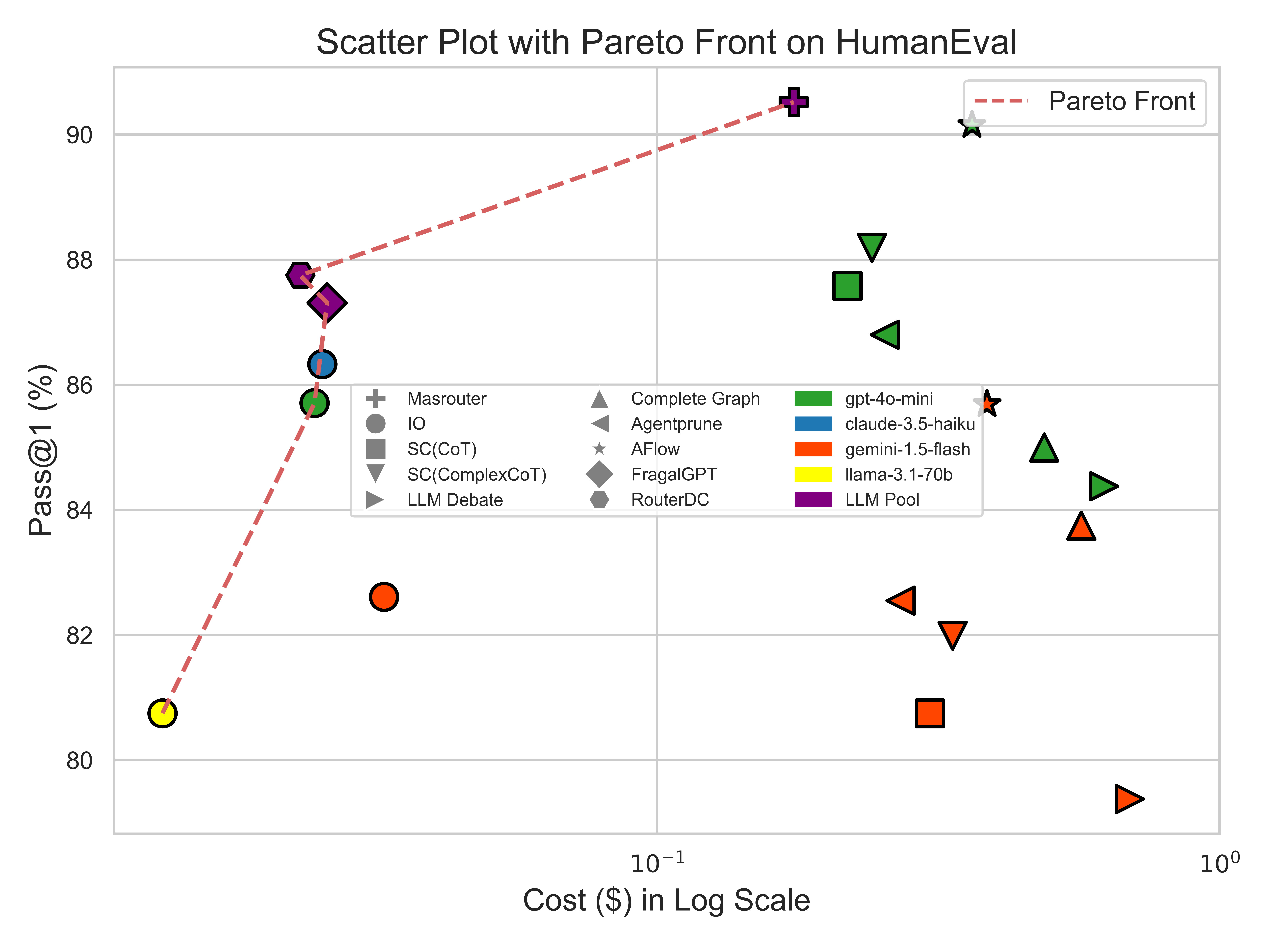}
\vspace{-2em}
\caption{The comparison of the performance and inference cost on the HumanEval dataset. Different shapes of the scatter points represent various types of baselines, while the different colors of the points indicate the use of different LLM backbones.}
\vspace{-0.2cm}
\label{fig:humanevalpareto}
\end{figure}

In this section, we present the specific overhead and performance of various baselines on the MBPP (\Cref{tab:mbppcost}) and HumanEval (\Cref{tab:humanevalcost}) datasets. The scatter plot with Pareto Front on HumanEval is shown in \Cref{fig:humanevalpareto}.
\begin{table*}[!t]
\centering
\begin{tabular}{cccc}
\hline
Method & LLM & Score(\%) & Cost(\$) \\
\hline
IO & \llmname{gpt-4o-mini} & 72.20 & 0.143\\
IO & \llmname{claude-3.5-haiku}  & 73.40 & 0.146\\
IO & \llmname{gemini-1.5-flash}  & 73.00 & 0.157\\
IO & \llmname{llama-3.1-70b} & 68.20 & 0.105\\
SC(CoT) & \llmname{gpt-4o-mini} & 73.00 & 0.449\\
SC(CoT) & \llmname{gemini-1.5-flash} & 72.00 & 0.548\\
SC(ComplexCoT) & \llmname{gpt-4o-mini} & 75.60 & 0.487\\
SC(ComplexCoT) & \llmname{gemini-1.5-flash} & 73.60 & 0.633\\
LLM Debate & \llmname{gpt-4o-mini} & 73.60 & 4.427\\
LLM Debate & \llmname{gemini-1.5-flash} & 73.40 & 4.529\\
Macnet(Complete Graph) & \llmname{gpt-4o-mini} & 75.20 & 2.932\\ 
Macnet(Complete Graph) & \llmname{gemini-1.5-flash} & 74.20 & 3.088\\
Agentprune & \llmname{gpt-4o-mini} & 75.00 & 1.215\\
Agentprune & \llmname{gemini-1.5-flash} & 75.60 & 1.352\\
AFlow & \llmname{gpt-4o-mini} & 82.20 & 1.723\\
AFlow & \llmname{gemini-1.5-flash} & 76.00 & 1.832\\
FragalGPT & llm pool & 74.40 & 0.139\\
RouterDC & llm pool & 75.20 & 0.145\\
\ourmethod & llm pool & 84.00 & 1.039 \\
\hline
\end{tabular}
\caption{Inference Cost-Performance on MBPP Dataset}
\label{tab:mbppcost}
\end{table*}

\begin{table*}[!t]
\centering
\begin{tabular}{cccc}
\hline
Method & LLM & Score(\%) & Cost(\$) \\
\hline
IO & \llmname{gpt-4o-mini} & 85.71 & 0.025\\
IO & \llmname{claude-3.5-haiku}  & 86.33 & 0.025\\
IO & \llmname{gemini-1.5-flash}  & 82.61 & 0.032\\
IO & \llmname{llama-3.1-70b} & 80.75 & 0.013\\
SC(CoT) & \llmname{gpt-4o-mini} & 87.58 & 0.218\\
SC(CoT) & \llmname{gemini-1.5-flash} & 80.75 & 0.306\\
SC(ComplexCoT) & \llmname{gpt-4o-mini} & 88.19 & 0.241\\
SC(ComplexCoT) & \llmname{gemini-1.5-flash} & 81.99 & 0.335\\
LLM Debate & \llmname{gpt-4o-mini} & 84.38 & 0.624\\
LLM Debate & \llmname{gemini-1.5-flash} & 79.38 & 0.693\\
Macnet(Complete Graph) & \llmname{gpt-4o-mini} &85.00 & 0.488\\
Macnet(Complete Graph) & \llmname{gemini-1.5-flash} & 83.75 & 0.568\\
Agentprune & \llmname{gpt-4o-mini} & 86.80 & 0.254\\
Agentprune & \llmname{gemini-1.5-flash} & 82.55 & 0.271\\
AFlow & \llmname{gpt-4o-mini} & 90.15 & 0.363\\
AFlow & \llmname{gemini-1.5-flash} & 85.69 & 0.386\\
FragalGPT & llm pool & 87.31 & 0.026\\
RouterDC & llm pool & 87.75 & 0.023\\
\ourmethod & llm pool & 90.52 & 0.185 \\
\hline
\end{tabular}
\caption{Inference Cost-Performance on HumanEval Dataset}
\label{tab:humanevalcost}
\end{table*}

\subsection{Training Cost}
In \Cref{tab:trainingcost}, we compare the training overhead of the SOTA methods that require training with \ourmethod on MATH and MMLU.
\begin{table*}[!ht]
\centering
\setlength\tabcolsep{3pt}
\begin{tabular}{lcccccc}
\Xhline{1.2pt} 
Method & \multicolumn{3}{c}{\textbf{MATH}
} & \multicolumn{3}{c}{\textbf{MMLU}}    \\
\cmidrule(lr){2-4} \cmidrule(lr){5-7}
  & {\makecell{Prompt\\token}} & {\makecell{Completion\\token}}  & {\makecell{Total\\cost (\$)}}  & {\makecell{Prompt\\token}} & {\makecell{Completion\\token}}  & {\makecell{Total\\cost (\$)}}\\
\hline
GPTSwarm & $23,031,287$ & $6,943,173$ & $7.63\$$&  $15,525,155$ & $3,983,745$ & $4.70\$$\\
AFlow & $321,813,314$ & $28,083,445$ & $21.75\$$  & $13,085,019$ & $11,239,502$  & $8.67\$$\\
\hline
\ourmethod & \cellcolor{gray!25}$3,235,288$ & \cellcolor{gray!25}$2,499,530$ & \cellcolor{gray!25}$3.56\$$ & \cellcolor{gray!25}$4,459,674$ & \cellcolor{gray!25}$2,904,656$ & \cellcolor{gray!25}$1.43\$$\\
\Xhline{1.2pt}
\end{tabular}
\caption{Training Cost comparison between \ourmethod and state-of-the-art baselines on MATH and MMLU.}
\label{tab:trainingcost}
\vspace{-0.9em}
\end{table*}

\section{The Module Profile}\label{app:profile}
In this section, we present the profiles of each module. We generate the LLM profile following GraphRouter~\citep{feng2024graphroutergraphbasedrouterllm} and construct the role pool following the method of Macnet~\citep{qian2024scaling}.
We have selected three distinct roles for each task, as presented in the ``LLM Profile'' and ``Role Profile'' boxes.

\clearpage
\begin{onecolumn}
\begin{tcolorbox}[notitle, sharp corners, breakable, 
     colframe=Periwinkle, colback=white, 
       boxrule=3pt, boxsep=0.5pt, enhanced, 
       shadow={3pt}{-3pt}{0pt}{opacity=1},
       title={LLM Profile},]
       \footnotesize
       {\fontfamily{pcr}\selectfont
\begin{lstlisting}[breaklines=true,showstringspaces=false]
llm_profile = [ 
{
    'Name': 'gpt-4o-mini',
    'Description': 'GPT-4o Mini is a smaller version of the GPT-4o language model, designed for faster inference and reduced memory usage. It retains the same capabilities as the full-size model, but with fewer parameters.
    The model costs $0.15 per million input tokens and $0.6 per million output tokens.
    In General Q&A Benchmark MMLU, GPT-4o-mini achieves an accuracy of 77.8.
    In Reasoning Benchmark GPQA, GPT-4o-mini achieves an accuracy of 40.2.
    In Coding Benchmark HumanEval, GPT-4o-mini achieves an accuracy of 85.7.
    In Math Benchmark MATH, GPT-4o-mini achieves an accuracy of 66.09.'
},
{
    'Name': 'claude-3-5-haiku-20241022',
    'Description': 'The new Claude 3.5 Haiku combines rapid response times with improved reasoning capabilities, making it ideal for tasks that require both speed and intelligence. Claude 3.5 Haiku improves on its predecessor and matches the performance of Claude 3 Opus.
    The model costs $0.1 per million input tokens and $0.5 per million output tokens.
    In General Q&A Benchmark MMLU, claude-3-5-haiku achieves an accuracy of 67.9.
    In Reasoning Benchmark GPQA, claude-3-5-haiku achieves an accuracy of 41.6.
    In Coding Benchmark HumanEval, claude-3-5-haiku achieves an accuracy of 86.3.
    In Math Benchmark MATH, claude-3-5-haiku achieves an accuracy of 65.9.'
},
{
    'Name': 'gemini-1.5-flash-latest',
    'Description': 'Gemini 1.5 Flash was purpose-built as our fastest, most cost-efficient model yet for high volume tasks, at scale, to address developers feedback asking for lower latency and cost.
    The model costs $0.15 per million input tokens and $0.6 per million output tokens.
    In General Q&A Benchmark MMLU, gemini-1.5-flash achieves an accuracy of 80.0.
    In Reasoning Benchmark GPQA, gemini-1.5-flash achieves an accuracy of 39.5.
    In Coding Benchmark HumanEval, gemini-1.5-flash achieves an accuracy of 82.6.
    In Math Benchmark MATH, gemini-1.5-flash achieves an accuracy of 74.4.'
},
{
    'Name': 'Meta-Llama-3.1-70B-Instruct',
    'Description': 'The Meta Llama 3.1 multilingual large language model (LLM) is a pretrained and instruction tuned generative model in 70B (text in/text out).
    The model costs $0.2 per million input tokens and $0.2 per million output tokens.
    In General Q&A Benchmark MMLU, Llama 3.1 achieves an accuracy of 79.1.
    In Reasoning Benchmark GPQA, Llama 3.1 achieves an accuracy of 46.7.
    In Coding Benchmark HumanEval, Llama 3.1 achieves an accuracy of 80.7.
    In Math Benchmark MATH, Llama 3.1 achieves an accuracy of 60.3.'
},
{
    'Name': 'deepseek-chat',
    'Description': 'DeepSeek-V3 is a cutting-edge, large-scale language model designed for advanced natural language processing (NLP) tasks.
    The model costs $0.27 per million input tokens and $1.1 per million output tokens.
    In General Q&A Benchmark MMLU, deepseek achieves an accuracy of 88.5.
    In Reasoning Benchmark GPQA, deepseek achieves an accuracy of 59.1.
    In Coding Benchmark HumanEval, deepseek achieves an accuracy of 88.4.
    In Math Benchmark MATH, deepseek achieves an accuracy of 85.1'
},
]
\end{lstlisting}
}
\end{tcolorbox}

\begin{tcolorbox}[notitle, sharp corners, breakable, 
     colframe=Periwinkle, colback=white, 
       boxrule=3pt, boxsep=0.5pt, enhanced, 
       shadow={3pt}{-3pt}{0pt}{opacity=1},
       title={Role Profile},]
       \footnotesize
       {\fontfamily{pcr}\selectfont
\begin{lstlisting}[breaklines=true,showstringspaces=false,label={box:role-profile}]
Math_role_profile = [
{
    "Name": "MathAnalyst",
    "MessageAggregation": "Normal",
    "Description": "You are a mathematical analyst. You will be given a math problem, analysis and code from other agents. 
    You need to first analyze the problem solving process, where the variables are represented by letters. 
    Then you substitute the values into the analysis process to perform calculations and get the results.",
    "OutputFormat": "Calculation",
    "PostProcess": "None",
    "PostDescription": "None",
    "PostOutputFormat": "None"
},
{
    "Name": "MathTeacher",
    "MessageAggregation": "PHP",
    "Description": "You are an excellent math teacher and always teach your students math problems correctly. 
    And I am one of your students.You will be given a math problem, teach me step by step how to solve the problem.",
    "OutputFormat": "Calculation",
    "PostProcess": "None",
    "PostDescription": "None",
    "PostOutputFormat": "None"
},
{
    "Name": "Inspector",
    "MessageAggregation": "Normal",
    "Description": "You are an Inspector. You will be given a math problem, analysis and code from other agents. 
    Check whether the logic/calculation of the problem solving and analysis process is correct(if present). 
    Check whether the code corresponds to the solution analysis(if present). 
    Give your own solving process step by step based on hints",
    "OutputFormat": "Answer",
    "PostProcess": "None",
    "PostDescription": "None",
    "PostOutputFormat": "None"
}
]
Coding_role_profile = [
{
    "Name": "Algorithm Designer",
    "MessageAggregation": "PythonInnerTest",
    "Description": "You are an algorithm designer. You will be given a function signature and its docstring by the user.
    You need to specify the specific design of the algorithm, including explanations of the algorithm, usage instructions, and API references.
    You can refer to specific examples.When the implementation logic is complex, you can give the pseudocode logic of the main algorithm.
    Your reply will be more concise.Preferably within fifty words.",
    "OutputFormat": "Text",
    "PostProcess": "None",
    "PostDescription": "None",
    "PostOutputFormat": "None"
},
{
    "Name": "BugFixer",
    "Description": "You are a programming expert. You will be given a function signature and its docstring by the user. Use a Python code block to write your full implementation (restate the function signature).",
    "OutputFormat": "CodeCompletion",
    "PostProcess": "PythonInnerTest",
    "PostDescription": "You need to provide modified and improved python code based on the current code implementation and problems that arise during testing.
    You can refer to specific examples.Write your full implementation (restate the function signature). ",
    "PostOutputFormat": "CodeCompletion"
},
{
    "Name": "Test Analyst",
    "MessageAggregation": "PythonInnerTest",
    "Description": "You are a Test Analyst. You will be given a function signature and its docstring by the user.
    You need to provide problems in the current code or solution based on the test data and possible test feedback in the question.
    You need to provide additional special use cases, boundary conditions, etc. that should be paid attention to when writing code.
    You can point out any potential errors in the code.Your reply should be more concise.Preferably within fifty words.",
    "OutputFormat": "Text",
    "PostProcess": "None",
    "PostDescription": "You are a programming expert. You will be given a function signature and its docstring by the user.
    Give your own answers to problems that arise in other implementations.
    Use a Python code block to write your full implementation (restate the function signature).",
    "PostOutputFormat": "CodeCompletion"
}
]

Commensense_role_profile =[
{
    "Name": "Critic",
    "MessageAggregation": "Normal",
    "Description": "You are an excellent critic. Please point out potential issues in other agent's analysis point by point. Give your critical opinion. Finally give the final result",
    "OutputFormat": "Answer",
    "PostProcess": "None",
    "PostDescription": "None",
    "PostOutputFormat": "None"
},
{
    "Name": "WikiSearcher",
    "MessageAggregation": "Normal",
    "Description": "Please give several key entities that need to be searched in wikipedia to solve the problem. ",
    "OutputFormat": "Keys",
    "PostProcess": "Wiki",
    "PostDescription": "You are a knowlegable expert in question answering. Please answer the question based on the explanation of the question keywords obtained from the wikipedia search.",
    "PostOutputFormat": "Answer"
},
{
    "Name": "Historian",
    "MessageAggregation": "Normal",
    "Description": "You research and analyze cultural, economic, political, and social events in the past, collect data from primary sources and use it to develop theories about what happened during various periods of history.",
    "OutputFormat": "Answer",
    "PostProcess": "None",
    "PostDescription": "None",
    "PostOutputFormat": "None"
}
]
\end{lstlisting}

}
\end{tcolorbox}

\begin{tcolorbox}[notitle, sharp corners, breakable, 
     colframe=Periwinkle, colback=white, 
       boxrule=3pt, boxsep=0.5pt, enhanced, 
       shadow={3pt}{-3pt}{0pt}{opacity=1},
       title={Reasoning Profile},]
       \footnotesize
       {\fontfamily{pcr}\selectfont
\begin{lstlisting}[breaklines=true,showstringspaces=false]
reasoning_profile = [
{
    'Name': 'IO', 
    'Description': 'In single-agent IO reasoning, a single agent directly gives an output based on the input.'
},
{   'Name': 'CoT', 
    'Description': 'In single-agent CoT reasoning, a single agent reasons step-by-step to achieve a goal.'
},
{   'Name': 'Chain',
    'Description': 'In multi-agent chain reasoning, multiple agents sequentially reason and pass information in a chain-like manner.'
},
{   'Name': 'FullConnected', 
    'Description': 'In multi-agent full-graph reasoning, multiple agents reason collectively over the entire graph structure.'
},
{   'Name': 'Debate',
    'Description': 'In multi-agent debate reasoning, multiple agents engage in a structured argumentative dialogue to explore different perspectives, challenge assumptions, and reach a consensus.'
},
{   'Name': 'Reflection', 
    'Description': 'In multi-agent reflection reasoning, multiple agents reflect on their own reasoning processes and outcomes to improve their performance.'
},
]
\end{lstlisting}
}
\end{tcolorbox}
\end{onecolumn}

\end{document}